\documentclass[journal]{IEEEtran}


\usepackage{stfloats}
\usepackage{times}
\usepackage{epsfig}
\usepackage{graphicx}
\usepackage{amsmath}
\usepackage{amssymb}
\usepackage{array}
\usepackage{tikz}
\usepackage{hyperref}

\begin{document}
%
\title{Deep Demosaicing\\for Polarimetric Filter Array Cameras}
%
%
%

\author{Mara Pistellato,
        Filippo Bergamasco,
        Tehreem Fatima,
        and~Andrea Torsello

\thanks{M. Pistellato, F. Bergamasco, T. Fatima and A. Torsello are with Department of Environmental Sciences, Informatics and Statistics (DAIS), Ca'Foscari University of Venice, 155 Via Torino, 30170 Venice, Italy. (email: mara.pistellato@unive.it, filippo.bergamasco@unive.it, tehreem.fatima@unive.it, andrea.torsello@unive.it)
}
}

\maketitle

\begin{abstract}
Polarisation Filter Array (PFA) cameras allow the analysis of light polarisation state in a simple and cost-effective manner. Such filter arrays work as the Bayer pattern for colour cameras, sharing similar advantages and drawbacks.
Among the others, the raw image must be demosaiced considering the local variations of the PFA and the characteristics of the imaged scene.

Non-linear effects, like the cross-talk among neighbouring pixels, are difficult to explicitly model and suggest the potential advantage of a data-driven learning approach. However, the PFA cannot be removed from the sensor, making it difficult to acquire the ground-truth polarization state for training.

In this work we propose a novel CNN-based model which directly demosaics the raw camera image to a per-pixel Stokes vector. Our contribution is twofold. First, we propose a network architecture composed by a sequence of \emph{Mosaiced Convolutions} operating coherently with the local arrangement of the different filters. Second, we introduce a new method, employing a consumer LCD screen, to effectively acquire real-world data for training. The process is designed to be invariant by monitor gamma and external lighting conditions.
We extensively compared our method against algorithmic and learning-based demosaicing techniques, obtaining a consistently lower error especially in terms of polarisation angle.

\end{abstract}


%
\IEEEpeerreviewmaketitle

\section{Introduction}

\IEEEPARstart{P}{olarimetric} imaging applications take advantage on the polarisation property of light to devise information on the acquired surface.
Indeed, reflected light exhibits a polarisation angle that can be directly related to surface material and normal, giving additional information on the captured scene.
The potential of such information makes this technology increasingly used in a wide range of modern Computer Vision tasks.
A popular application is Shape from Polarisation, where per-pixel surface normals are retrieved starting from polarisation cues. Such approaches range from classical physics-based methods \cite{atkinson2006recovery, yu2017shape} to Deep Learning approaches \cite{ba2019physics}.
Several methods extract depth information by pairing polarimetric data with other priors, like lighting constraints \cite{smith2018height} or coarse depth maps \cite{kadambi2017depth}.

\begin{figure}[h]
    \centering
    \includegraphics[width=0.99\linewidth]{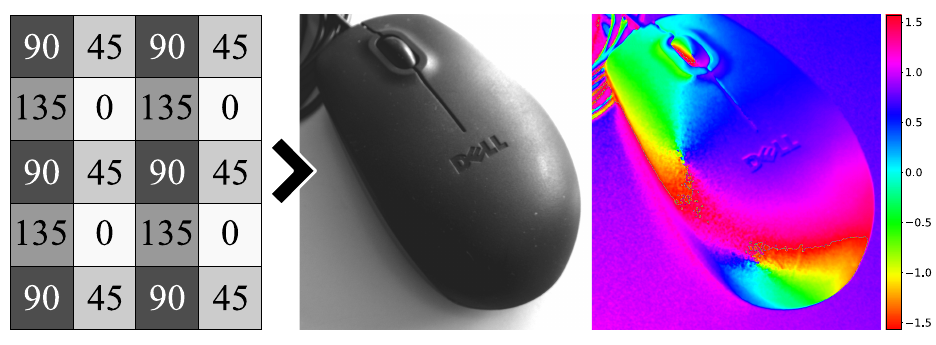}
    \caption{Our method demosaics a PFA camera image to produce a high-resolution intensity image (center) and angle of polarisation (right).}
    \label{fig:banner}
\end{figure}

Recent works propose the fusion of polarisation imaging with classical geometrical techniques for 3D reconstruction, like multi-view stereo \cite{cui2017polarimetric, zhao2020polarimetric}, polarimetric relative pose estimation \cite{cui2019polarimetric}, dense monocular SLAM \cite{yang2018polarimetric} or three-view geometry \cite{chen2018polarimetric}.
Other applications include material classification \cite{chen1998polarization}, transparent objects analysis \cite{miyazaki2007shape}, radiometric calibration \cite{teo2018self} or facial reconstruction \cite{ghosh2011multiview}.

Typically, polarimetric images can be captured using either a linear polariser rotating in front of the lenses, or with  PFA cameras. The latter are fabricated by superimposing a linearly-polarised filter array to the underlying focal-plane sensor array. In this way, the sensor grid is divided into $2 \times 2$ macro-pixels with a repeated pattern of filters oriented at $0^\circ, 45^\circ, 135^\circ, 90^\circ$ (Figure~\ref{fig:banner}, left). The advantage of this approach is that we can capture four different orientations with a single shot, allowing the recovery of both intensity and linear polarisation state for each macro-pixel. This in general simplify the acquisition process, in particular for moving objects.

On the other hand, this setup makes more difficult to obtain the full-resolution polarimetric image for two reasons.
First, macro-pixels structure trades spatial resolution with the ability to recover the light beam polarisation. This is conceptually identical of what happens with the Bayern pattern in colour imaging, that is, the image must be demosaiced to unravel the filter pattern into a $4$-channel polarised image.
The second problem is that each individual pixel is subject to different sources of noise and a possible cross-talk effect among neighbouring pixels~\cite{Gimenez2020}.
The former is typically assumed to depend on the acquired image (ie. how to interpolate across the edges?) and the latter on the specific camera instance (ie. how to equalize each array cell?). The result is the difficulty to conceive a purely algorithmic approach taking into account all the aforementioned factors.

Recently, learning-based demosaicing techniques have been explored both for colour and polarimetric cameras. Considering their popularity, CNNs are the primary candidate for this task but their application is not as straightforward as it might seem.
As a matter of fact, Deep Learning requires a huge amount of data to be effective, but such data are difficult to obtain because the filter cannot be removed from the sensor. For this reason, almost all the existing approaches use surrogate data, either completely synthetic or captured from cameras with different technologies (like rotating filters in front of the lenses).
We argue that, in this way, the model will never learn the non-linear interactions among pixels in PFAs that deemed the learning approach to be effective in the first place.

In this paper, we first describe a practical data acquisition process employing an LCD screen that takes into account the monitor properties and allows for the creation of a real-world demosaicing dataset for a specific camera.
Then, we present a CNN architecture called PFADN (PFA Demosaicing Network) explicitly designed for this task\footnote{Source code is available at \url{https://github.com/DAISCVprojects/PFADN}}. It involves a novel \textit{Mosaiced Convolution block}, devised to learn different kernels for each four different arrangements of the filter orientations in the mosaiced image.
Our model outputs both the full-resolution demosaiced intensity image and the polarisation angle values, outperforming the state-of-the-art algorithmic and learning-based methods as well. The advantage is that it can be trained on a specific kind of camera, simultaneously accounting for the imaged scene and the cross-talking effects produced at a sensor level.

\section{Related Work}

Demosaicing is essentially an interpolation problem, producing a full-resolution image from a sparse set of samples capturing different characteristics of the light (intensity at certain wavelengths or polarization).
The most popular is Colour demosaicing, designed to interpolate images acquired with Colour Filter Array (CFA) cameras. This is a widely covered topic in the literature, because the Bayer pattern is available since 1970s.
A good survey is given in \cite{li2008image}, with a comparative study of 11 methods tested in different conditions. In general, interpolation can be performed in spatial or frequency domain. The former works directly on image pixel, and are typically based on a sequential approach in which the luminance is reconstructed first and chrominance is then obtained based on the recovered luminance. The latter are designed as anti-aliasing filters in the frequency domain (Fourier~\cite{Fourier2005} or wavelet~\cite{Wavelet2004}). A more recent survey is provided by \cite{Mihoubi2018}.

Not all the methods designed for colour images can be easily ported to a PFA camera because the redundancy of the green colour (twice more pixels than the others) is usually exploited to drive an edge-directed interpolation~\cite{Lu2003}.
For this reason, several ad-hoc solutions have been developed for polarimetric demosaicing. Basic techniques propose bilinear and bicubic interpolation \cite{ratliff2009interpolation, ratliff2009interpolation, gao2011bilinear} operating isotropically at each image pixel.
To compensate for edge artefacts, more sophisticated techniques take into account the intensity of the underlying image. For instance, \cite{zhang2016image} uses the correlation of intensity measures at different orientations to drive the interpolation only along, and not across, the edges.
Similarly, \cite{gao2013gradient} explicitly computes the gradient image to expand bicubic and bilinear interpolation to incorporate gradient selectivity features. Finally, \cite{ahmed2017residual} computes the difference between an observed and a tentatively estimated pixel value to refine the interpolation.

PFA cameras are affected by several sources of error: some are common to all silicon sensor devices (dark-current, salt-and-pepper, etc.), others are due to imperfections in the manufacturing of PFA itself. Indeed, each individual filter exhibits slight variations in transmission, diattenuation and orientation with respect to the expected angles~\cite{Bass2009}.
To account for these factors, several algorithms have been proposed in the recent past~\cite{gimenez2020calibration, powell2013calibration, chen2015calibration} but they usually employ a specialised hardware setup providing a perfectly (controlled) polarised light source. This suggests that a proper demosaicing technique should ideally account not only the image edges but also the different non-linear cross-talking effects that may occur between pixels.

For this reason, some recent papers cast polarimetric demosaicing into Deep Learning domain.
First attempts were described by \cite{zhang2018learning, zeng2019end}, with an end-to-end architecture designed to take the mosaiced image as input and produce demosaiced intensity, angle and Degree of Polarisation images as outputs.
Differently from our approach, such architectures do not consider the underlying mosaic pattern: the same convolution kernels are applied uniformly (with unitary stride) among pixels, mixing different orientation patterns. We will show in the experimental section that this approach is far from being optimal.
Other recent learning-based approaches for polarimetric demosaicing are proposed in \cite{sargent2020conditional,song2021transcending}.

Another effective way to demosaic a PFA camera is to simply recover the Stokes vector for each macropixel, and then upscale the result to obtain a full-resolution image.
Also in this context, the state-of-the-art is represented by CNN-based approaches~\cite{Dong2016,Kim_2016_CVPR,Zhang_2018_ECCV}.
Other techniques involve pixel-level fusion \cite{kwan2018demosaicing}, combining results from other debayering algorithms.
However, all such methods are designed to interpolate image intensities (and colours) and tend to behave poorly with the angle of polarisation.
Finally, Wen et al. \cite{wen2019joint} propose joint chromatic and polarimetric demosaicing, acting as a bridge between classic colour camera and PFA camera demosaicing methods.



\section{PFA Camera Demosaicing}

Polarisation is a basic property of light originating from its vectorial nature. Indeed, the optical field is composed by two orthogonal sinusoidal components $E_x, E_y$, oscillating in the plane transverse to the direction of propagation. At a certain instant of time, the locus of points described by such components is algebraically represented by an ellipse conveying the underlying polarisation state \cite{goldstein2017polarized}.
For Computer Vision applications, we are particularly interested in degenerate states in which the light is either linearly-polarised (ie. the ellipse minor axis is zero) or circularly-polarised (ie. the ellipse is a circumference).

It is common to describe the polarisation state with a vector $\begin{pmatrix} S_0 & S_1 & S_2 & S_3\end{pmatrix}$ of the so-called Stokes Parameters. In this model, $S_0$ quantifies the light intensity, as acquired by ubiquitous grayscale cameras. $S_1$ and $S_2$ convey information on the linear polarisation, measuring the difference between vertical-horizontal and $45^\circ$-$135^\circ$ components respectively. Finally, $S_3$ weights the preponderance of clockwise over counterclockwise circular polarisation.

Filter orientations on PFA cameras are chosen to easily obtain the first three Stokes parameters.
Assuming a scene with no circular polarisation, for each macro-pixel the measured intensities $I_0, I_{45}, I_{135}, I_{90}$ can be used to derive the Degree of Linear Polarization (DoLP) (\ref{eqn:DOLP}) and the Angle of Linear Polarisation (AoLP) (\ref{eqn:AOLP}):


\begin{eqnarray}
\mbox{DoLP}(S_0,S_1,S_2) &=& \frac{\sqrt{S_1^2+S_2^2}}{S_0} \label{eqn:DOLP} \\
\mbox{AoLP}(S_1,S_2) &=& \frac{1}{2}\arctan{\frac{S_2}{S_1}} \label{eqn:AOLP}\\
S_0 &=& I_0 + I_{90} \nonumber \\
S_1 &=& I_0 - I_{90} \nonumber \\
S_2 &=& I_{45} - I_{135} = S_0 - 2 I_{135} \nonumber
\end{eqnarray}

We consider our PFA camera as a non-linear function $\mathcal{F}(I_\mathcal{P}) \to (I, \Phi)$ mapping the input (mosaiced) image $I_\mathcal{P} \in \Omega_{M \times N}$ to the output (demosaiced) intensity image $I = S_0 \in \Omega_{M \times N}$ and polarisation angle $\Phi = \mbox{AoLP}(S_1,S_2) \in [-\frac{\pi}{2} \ldots \frac{\pi}{2}]_{M \times N}$. We model $\mathcal{F}$ as a Convolutional Neural Network (§~\ref{sec:architecture}) whose weights are estimated a supervised learning process. To account for the huge amount of data needed to generalize, an initial training is performed with purely synthetic data as would have been generated by an ideal polarised filter array under common scene conditions. Then, such generic model is refined to a specific camera instance to learn the local non-linear effects among neighbouring pixels. For the first time, this operation is performed with real-world data acquired by the PFA device itself with a novel technique described in the next section.

\subsection{Acquiring the Training Data}\label{sec:calibration}

\begin{figure*}[t]
\begin{center}
\includegraphics[width=0.9\linewidth]{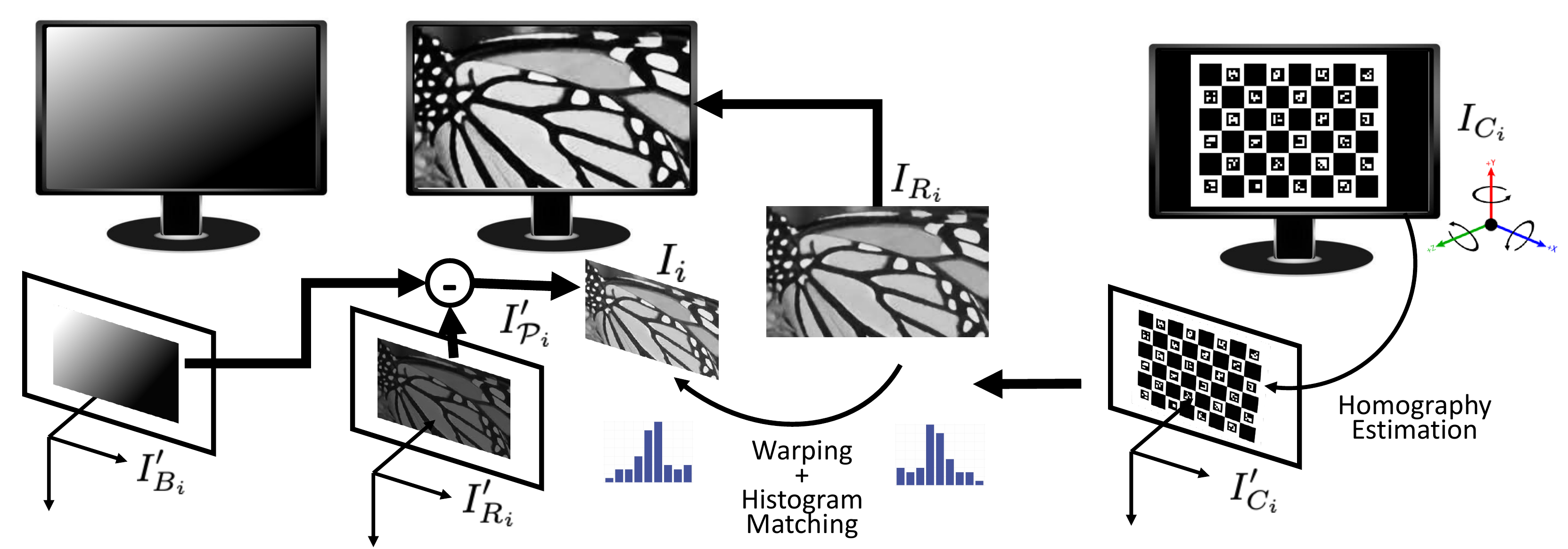}
\end{center}
   \caption{Overview of the proposed data aquisition procedure. An LCD screen is observed in several different poses. Three images are acquired for each pose: (i) a Charuco board $I_{C_i}$, (ii) a random sample from the DIV2K dataset $I_{R_i}$ and (iii) a black image. The black image is subtracted to $I_{R_i}$ to remove ambient illumination. The Charuco board is used to estimate the homography mapping $I_{R_i}$ to the camera image plane. Additionally, the non-linear intensity response of the screen is removed via Histogram Matching. The resulting image $I_i$ represent the ideal $I_{R_i}$ as if it had been acquired with the same camera but without the PFA. $I_i$ and the ground truth polarization angle $\Phi_i$ (computed from the homography) are the expected output of the model.}
\label{fig:DataAcquisition}
\end{figure*}

Since our model is to be refined for a specific camera instance, the supervised learning process requires several real-world input-output instances $\big(I_{\mathcal{P}_i}, (I_i, \Phi_i)\big)$. However, the filter array cannot be removed nor manually rotated so it is not trivial how to obtain the true $I$ and $\Phi$ corresponding to the observed mosaiced input $I_\mathcal{P}$.
This operation is generally referred as PFA calibration, but it relies on complex hardware setups to provide light stimuli with known Stokes vectors to the camera~\cite{Gimenez2020}. Albeit being expensive, such techniques cannot provide, in the same image, variable light intensity and polarisation to the camera pixels. Consequently, we cannot ``learn to demosaic'' since every $I_i$ will show no edge and no intensity variations.

\begin{figure*}[t]
\begin{center}
\includegraphics[width=0.9\linewidth]{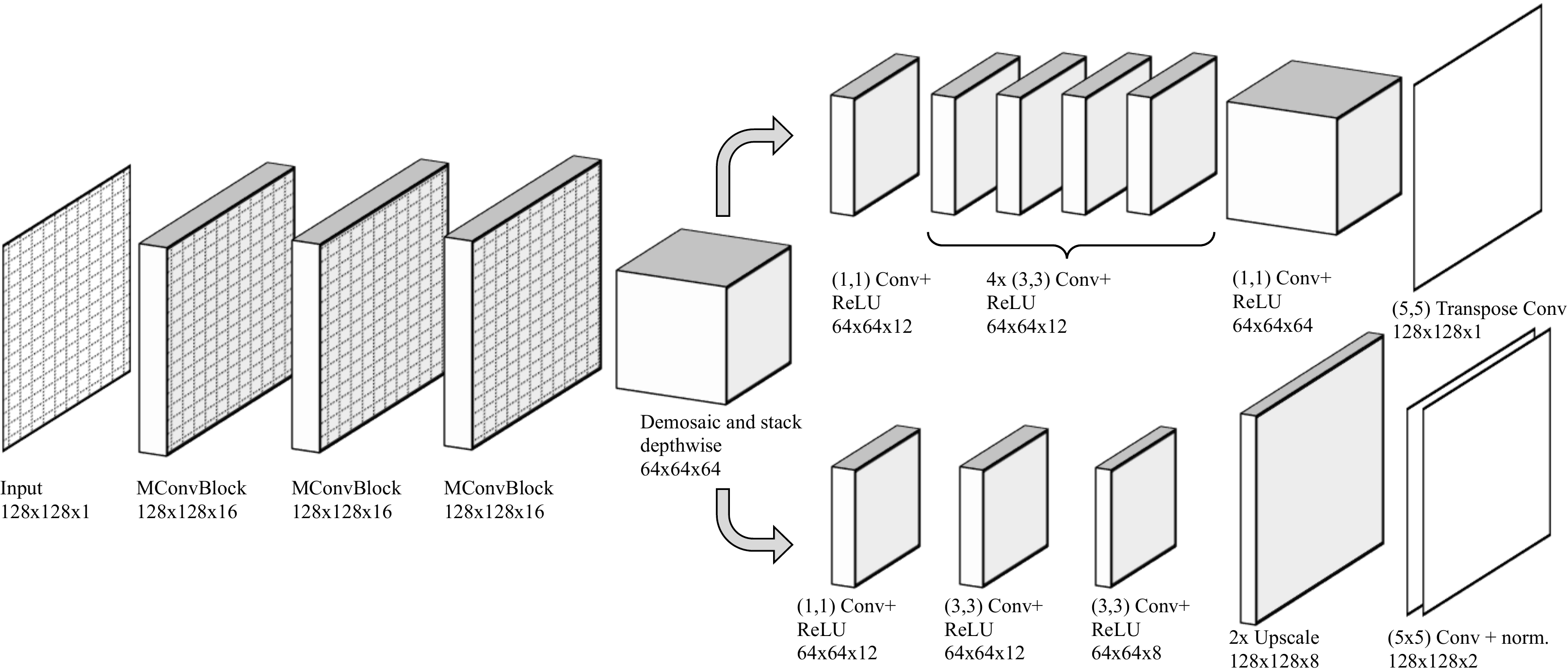}
\end{center}
   \caption{Our proposed PFADN architecture. The initial feature extraction is performed by $3$ mosaiced convolution blocks (§\ref{sec:mconv}). Then, pixels are demosaiced and each polarisation angle is stacked depthwise to produce a $64 \times 64 \times 64$ cube. Henceforth, network splits in two branches. The first produces the intensity image $I$ and the second the vector components of twice the angle of linear polarisation $\Phi$.}
\label{fig:PFADN}
\end{figure*}

Recently, \cite{Wang2019} proposed to use a consumer LCD screen as a source of (partially controllable) polarised light. The idea is promising, and we also embrace this approach. However, their method cannot be used as-is because the following factors should be properly accounted:
\begin{enumerate}
    \item There is a non-linear relationship between the monitor input image intensity and the produced output. The well-known \emph{Gamma} function is usually a good approximation for that, with values ranging from $2.1$ to $2.3$. Authors of \cite{Wang2019} stated that $2.2$ is a good value for most of the cases, but we argue that this is a too simplistic assumption and we will end up learning to correct the monitor gamma instead of the camera pixels' non-linear effects.
    \item The observed angle of polarisation depends on the monitor pose $(\mathbf{R},t)$, but it is not constant across the extent of the screen due to perspective distortion. Polarisation angle at each pixel is dominated by the Yaw angle of the rotation $\mathbf{R}$ but only if the other two are almost zero. Moreover, light passing through a \emph{tilted-polariser} is subject to non-trivial interactions that should (at least partially) be modelled~\cite{Korger13}.
    \item Light emitted by an LCD screen is linearly polarised with an unknown angle $\alpha$ with respect to the monitor pixel grid. Most of the times, such angle is either $0^\circ, 90^\circ$ or $45^\circ$ but the fabrication process does not guarantee the precision of such value, caring only to the relative orientation between the liquid crystals and the outermost polariser. In other words, $\alpha$ should be estimated as well.
    \item Part of the light emitted by the screen is due to ambient reflections. Unfortunately, such reflections are also linearly polarised but with a different angle, depending on the screen surface normal. Such component should be removed to avoid introducing biases in the angle estimation.
\end{enumerate}

Training data are acquired as follows. First, the camera is mounted on a tripod
featuring an adjustable joint to rotate the camera (at least $90^\circ$) around
its optical axis. This is quite common for commercial tripods to allow taking
pictures both in portrait and landscape\footnote{Alternatively, one can use an
LCD screen that can be freely rotated clockwise or counterclockwise}. Camera is
placed in front of an LCD screen in several different poses. In each pose the
screen is roughly perpendicular to the optical axis but rotated with any angle
around it (yaw angle). Note that it is not important for the screen to be
``exactly" perpendicular since the following operations will take care of it.
For each pose $i=1 \ldots K$, we display three images on the screen captured in
sequence with the camera. The first is a ChArUco
board~\cite{Aruco2014,DeepCharuco2019} $I_{C_i}$ designed to fill the entire
screen. The second is a random image $I_{R_i}$ from the DIV2K
dataset~\cite{Timofte_2018}, and the third image $I_{B_i}$ is completely black.
We denote with $I_{C_i}', I_{R_i}'$ and $I_{B_i}'$ the corresponding (mosaiced)
images acquired by the camera.

The $i^{th}$ input training sample is computed by subtracting the black image
to the random  image: $I_{\mathcal{P}_i}' = I_{R_i}' - I_{B_i}'$. This will remove
most of the unwanted (linearly polarised) reflections from the surrounding
environment. We also naively demosaic this image to compute the first three
Stoke parameters $S_{0_i}, S_{1_i}, S_{2_i}$ for each $2 \times 2$
macro-block\footnote{The resulting naively demosaiced polarisation images will
have $\frac{1}{4}$ the resolution of the camera input image}.  To compute the
corresponding network output $(I_i, \Phi_i)$ we take advantage of the acquired
ChArUco board to estimate the planar Homography $\mathbf{H}_i$ mapping points
from the LCD screen to the camera image space. Note that $\mathbf{H}_i$ can be
computed without calibrating the camera under the assumption that the lens
distortion is negligible. The output demosaiced intensity image $I_i$ is
obtained by first remapping $I_{R_i}$ with the homography $\mathbf{H}_i$ and then
by matching the resulting image histogram with the histogram of $S_{0_i}$.

Let's clarify this concept. $I_i$ is expected to have the same resolution of
$I_{\mathcal{P}_i}'$, representing the ideal $I_{R_i}$ as if it had been acquired
with the same camera with no PFA. Since we cannot remove the filter, we propose
to map the original displayed image $I_{R_i}$ to the camera image space as in
Augmented Reality applications. However, we must take into account that the
actual intensity response of the LCD screen is quite different from the input
signal. In particular, monitor gamma and other non-linear filters will
``enhance" the image in a quite unpredictable way. To fix that, we take
advantage of the corresponding demosaiced (and down-sampled) intensity image
$S_{0_i}$ to compute the intensity histogram $\mathcal{H}$ of the observed screen
area. Then, the histogram of $I_{R_i}$ remapped with $\mathbf{H}_i$ is matched to
$\mathcal{H}$ to get a realistic response regardless the non-linear
transformations introduced by the screen. Note that, since the intensity
histogram discards spatial information, computing it on a down-scaled version
of the acquired image is not a limitation (we still have plenty of pixels to
compute a reliable histogram). The only caution is to choose an image $I_{R_i}$
with enough variability in the intensity values.

The output demosaiced AoLP $\Phi_i$ is tricky to obtain accurately since light
rays entering the camera are not orthogonal to the screen. Consequently, we
cannot just consider the Yaw angle of $\mathbf{R}$ as suggested in
\cite{Wang2019}. Discarding complex internal inter-reflections due to the
multi-layer nature of the LCD, we model the polarisation state of screen pixels
as a vector field lying on the screen outermost surface. Such field is composed
by contravariant parallel vectors oriented with angle $\alpha$, depending on
the manufacturer fabrication process. Geometrically, every screen point
$p=\begin{pmatrix}x & y & 1\end{pmatrix}^T$ is associated with a vector
$v=\begin{pmatrix} v_x & v_y & 0\end{pmatrix}^T=\begin{pmatrix} \cos \alpha &
\sin \alpha & 0 \end{pmatrix}^T$. When imaged, the vector field is projected
into the camera space through the Homography $\mathbf{H}_i$ resulting in points
$p'$ and vectors $v'$: \begin{eqnarray} p' &=& \mathcal{P}_{H_i}(p)\\ \hat{v}'
&=& \mbox{J}_\mathcal{P} ( p' )^{-T} v \nonumber \\ v' &=& \hat{v}' / \|
    \hat{v}' \| \\ \mathcal{P}_H(p) &=& \mathbf{H}_i p / (\mathbf{H}_{i_{(3,1)}}p)
    \label{eq:homfunction} \end{eqnarray} \noindent where $\mbox{J}_\mathcal{P}
    ( p' )$ is the Jacobian of the homography projection function
    (\ref{eq:homfunction}) evaluated in $p'$ and $\mathbf{H}_{i_{(3,1)}}$ is the
    last row of $\mathbf{H}_i$. Finally, the arctangent of $v'$ gives the AoLP
    $\Phi_i$ for each pixel. This operation is purely geometrical, and hence
    the resulting $\Phi_i$ can be computed at any desired resolution. Note
    that, this way, training data will comprise samples in which polarisation
    angles cannot assume any possible distribution because are bound to the
    planar nature of the target. However, angles in $\Phi_i$ still exhibit some
    variability in a perspective camera.

Since the angle $\alpha$ is unknown, we must estimate it beforehand. To do so,
a non-linear least squares optimization is performed so that the projected
vectors $v'$ are as close as possible to the vectors obtained by computing the
AoLP with the downsampled $S_{1_i}$ and $S_{2_i}$ as in (\ref{eqn:AOLP}).

\subsection{PFADN Architecture}\label{sec:architecture}

When enough data are acquired, we can proceed by training our PFA camera model.
We propose a CNN architecture as  sketched in Fig.~\ref{fig:PFADN}. Input is
composed by a $128 \times 128 \times 1$ mosaiced image with filters disposed in
$2 \times 2$ macroblocks with angles $90^\circ, 45^\circ, 135^\circ, 0^\circ$.
An initial sequence of $3$ \emph{Mosaiced Convolution Blocks}
(§~\ref{sec:mconv}) extracts features from the underlying input image. Each
block convolves the input tensor with a total of $64$ $2 \times 2 \times d$
kernels to produce a mosaiced output $\mathcal{T}$ with size $128 \times 128
\times 16$. From a functional point of view, a mosaiced convolution behaves
like the 2D convolutions found in standard CNNs, but taking advantage and
preserving the underlying polarisation pattern. This acts as a feature
extraction stage, augmenting the input image from $1$ to $16$ channels.

After features extraction, $\mathcal{T}$ is demosaiced into four $64 \times 64
\times 16$ feature tensors corresponding to the four filter orientations.  Such
tensors are stacked depth-wise to create a $64 \times 64 \times 64$ cube;
specifically the first $16$ channels are composed by extracting pixels at
coordinates $(2u, 2v)$ from $\mathcal{T}$. Channels $16 \ldots 31$ with pixels
at coordinates $(2u+1,2v)$, channels $32 \ldots 47$ with $(2u,2v+1)$ and
channels $48 \ldots 63$ with $(2u+1,2v+1)$.

At this point, the image is demosaiced but at $1/4$ of its original resolution.
Henceforth, the network splits in two parallel branches. The top branch
computes an intensity image $\hat{I}$, corresponding to the first Stoke
parameter $S_0$. This is a typical super-resolution problem, in which the CNN
should produce an up-scaled version of the input data based on what learned
during the training. We implemented the structure proposed in \cite{Dong2016}
from the \emph{Shrinking} step onward (the feature extraction is not necessary
as already performed by our MConv blocks). As suggested by the authors, the 2x
up-scaling is delayed until the final transposed convolution (or deconvolution)
with stride $2$ and filter size $(5,5)$.

The second branch deals with the angle of polarisation. However, angles are
harder to train if we do not account for their periodic nature. Instead of
directly producing the AoLP $\hat{\Phi}$, our network outputs a 2-channel image
$\mathcal{A}$ with vectors $a_{u,v} \in \mathbb{S}^1 = (\cos( 2
\hat{\Phi}_{u,v}), sin( 2 \hat{\Phi}_{u,v}))$ for each pixel $(u,v)$. By
doubling the angles, wrapping occurs every $2 \pi$ radians instead of $\pi$
\footnote{note that a polarisation angle of $\xi$ cannot be distinguished from
$\xi + k\pi$} enabling us to use the simple $L_2$ loss between the computed
angle vectors and the ones acquired from the monitor pose. This branch is
composed by a sequence of 2D convolutions with a decreasing number of kernels,
a non-trainable 2x upscale followed by a convolution to produce the output
$2$-channel tensor. Finally, values are projected to $\mathbb{S}^1$ by a
channel-wise normalization to ensure a unitary squared norm.  Since a camera
has a far greater resolution than the input tensor, the image is divided in a
uniform grid composed by cells measuring $128 \times 128$ pixels. The PFADN is
then repeatdly executed to demosaic each grid cell. The operation can be
computed in parallel, with a minimum impact on the final demosaicing
performance.

\subsection{Mosaiced Convolutions}\label{sec:mconv}

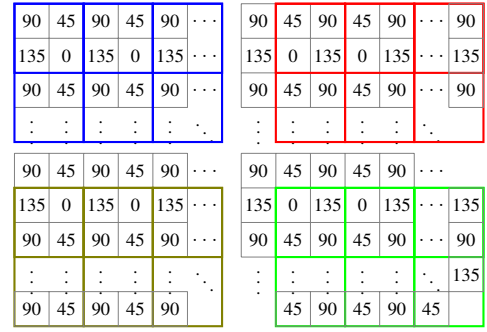
\begin{figure}[t]
\begin{center}
\begin{tikzpicture}[scale=0.46]
\let\clearpage\relax
\include{figures/drawing1}
\end{tikzpicture}
\begin{tikzpicture}[scale=0.46]
\let\clearpage\relax
\include{figures/drawing2}
\end{tikzpicture}\\
\begin{tikzpicture}[scale=0.46]
\let\clearpage\relax
\include{figures/drawing3}
\end{tikzpicture}
\begin{tikzpicture}[scale=0.46]
\let\clearpage\relax
\include{figures/drawing4}
\end{tikzpicture}
\end{center}
    \vspace{-0.3cm}
   \caption{The $4$ different convolution patterns used in our network. Top-left is generated by $\mbox{MConv}(T,0,0,d)$ (no cropping and no padding occurs). Top-right is generated by $\mbox{MConv}(T,0,1,d)$, implemented by cropping the first column and duplicating the last. Bottom-left is generated by $\mbox{MConv}(T,1,0,d)$ cropping the first row and duplicating the last. Bottom-right is produced by $\mbox{MConv}(T,1,1,d)$ cropping both the first row and column. In all the cases, a stride 2 convolution ensures that filter orientations remain consistent when swiping the filter along the tensor.}
   \vspace{-0.5cm}
\label{fig:mosaiced}
\end{figure}

It is very important to consider the PFA structure when doing the convolutions. For example, if we simply convolve the mosaiced image with a $3 \times 3$ kernel and unitary stride, the underlying orientation pattern will depend on the kernel location. Since the same kernel is swept along the entire image extent, is extremely difficult for the network to learn meaningful features from the data.
A solution is to use a kernel with size multiple of $2$ (PFA pattern is composed by $2 \times 2$ repeated tiles) and a stride equal to the kernel size. Since the stride shrinks the output, the alternative producing the minimum reduction is to use a $2 \times 2$ kernel with stride $2$ as shown in Fig.\ref{fig:mosaiced} (top-left).
This works, but is sub-optimal because each kernel will operate on the same macro-pixel and never across adjacent macro-pixels. Conceptually, this conveys the same information as unraveling each tile channel-wise and then convolving the resulting $4$-channel image.

If we restrict to $2 \times 2$ kernels, the $4$ possible patterns are summarized in Fig. \ref{fig:mosaiced}. Each one is obtained by shifting the image $1$ column to the left or $1$ row to the top. With this principle in mind, we define the $\mbox{MConv}(T,r,c,d)$ operator as follows:

\begin{enumerate}
    \item Crop the first $r$ rows and $c$ columns from the input $T$
    \item Duplicate the last $r$ rows and $c$ columns of tensor $T$
    \item Convolve $T$ with $d$ different kernels with size $2 \times 2$ and stride $2$. Then, apply Bias and ReLU.
    \item Upscale each output $2\times$
    \item Mask each output by tiling the pattern $M_{2r+c}$
\end{enumerate}

\noindent Where $M_0 = $\begin{tikzpicture}[scale=0.4]
\draw[step=1cm,gray,very thin] (0.0,0.0) grid (2,2);
\node[scale=0.6] at (0.5,1.5) {1};
\node[scale=0.6] at (1.5,1.5) {0};
\node[scale=0.6] at (0.5,0.5) {0};
\node[scale=0.6] at (1.5,0.5) {0};
\end{tikzpicture}, $M_1 = $\begin{tikzpicture}[scale=0.4]
\draw[step=1cm,gray,very thin] (0.0,0.0) grid (2,2);
\node[scale=0.6] at (0.5,1.5) {0};
\node[scale=0.6] at (1.5,1.5) {1};
\node[scale=0.6] at (0.5,0.5) {0};
\node[scale=0.6] at (1.5,0.5) {0};
\end{tikzpicture}, $M_2 = $\begin{tikzpicture}[scale=0.4]
\draw[step=1cm,gray,very thin] (0.0,0.0) grid (2,2);
\node[scale=0.6] at (0.5,1.5) {0};
\node[scale=0.6] at (1.5,1.5) {0};
\node[scale=0.6] at (0.5,0.5) {1};
\node[scale=0.6] at (1.5,0.5) {0};
\end{tikzpicture},$M_3 = $\begin{tikzpicture}[scale=0.4]
\draw[step=1cm,gray,very thin] (0.0,0.0) grid (2,2);
\node[scale=0.6] at (0.5,1.5) {0};
\node[scale=0.6] at (1.5,1.5) {0};
\node[scale=0.6] at (0.5,0.5) {0};
\node[scale=0.6] at (1.5,0.5) {1};
\end{tikzpicture}.

Then, the Mosaiced Convolution Block is implemented by summing the effect of the four Mosaiced Convolutions:
\begin{eqnarray*}
\mbox{MConvBlock}(T,d) = \mbox{MConv}(T,0,0,d) + \mbox{MConv}(T,1,0,d) \\
                       + \mbox{MConv}(T,0,1,d) + \mbox{MConv}(T,1,1,d).
\end{eqnarray*}

\noindent In practice, $\mbox{MConvBlock}(T,d)$ takes a mosaiced input tensor $T$ (with any depth) and produces a mosaiced output vector $T'$ with depth $d$. Width and height of the tensor and the underlying pattern structure are preserved so multiple MConvBlocks can be stacked as usual.

\begin{figure*}
    \centering
    \includegraphics[width=0.47\linewidth]{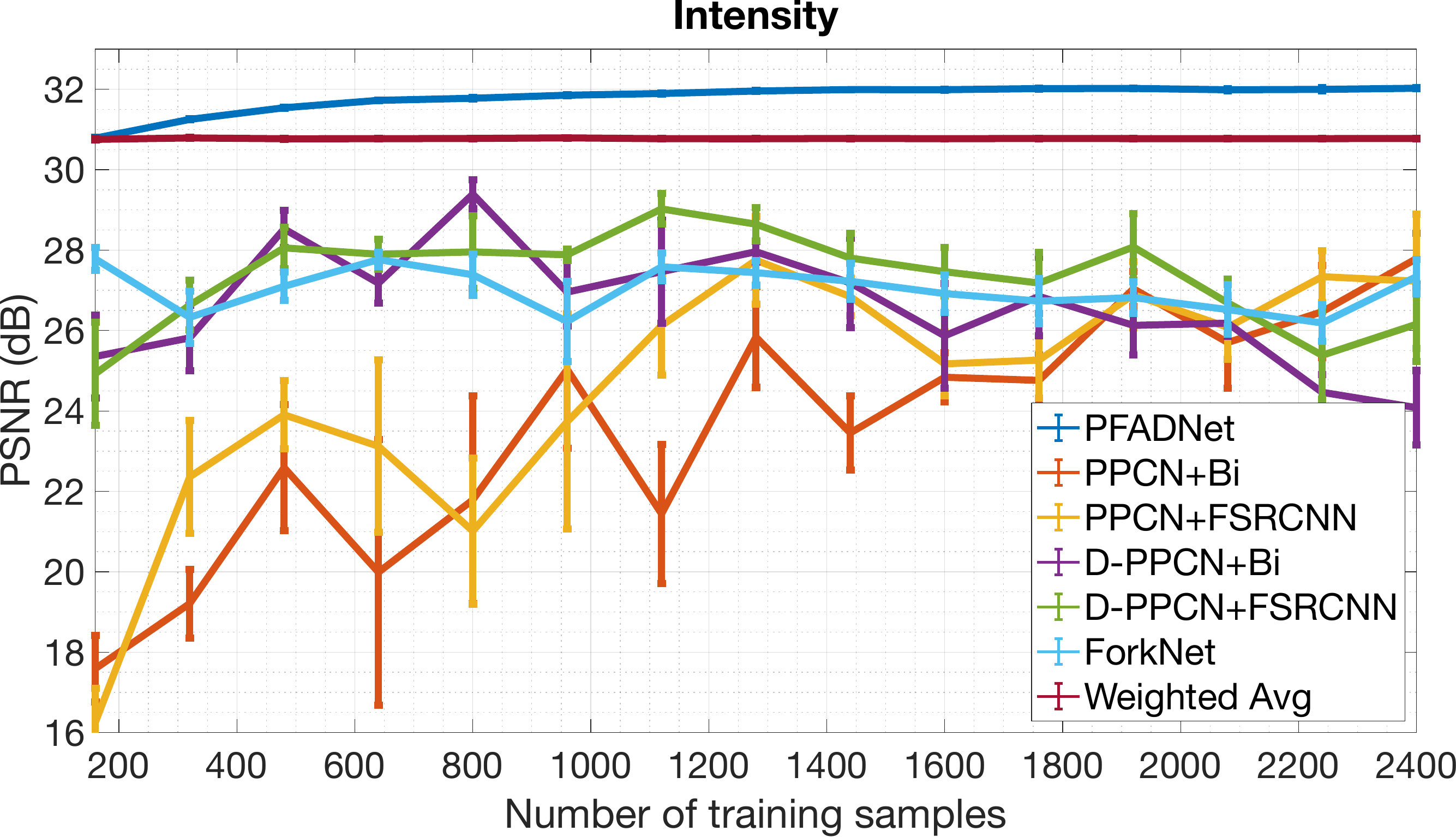}
    \includegraphics[width=0.47\linewidth]{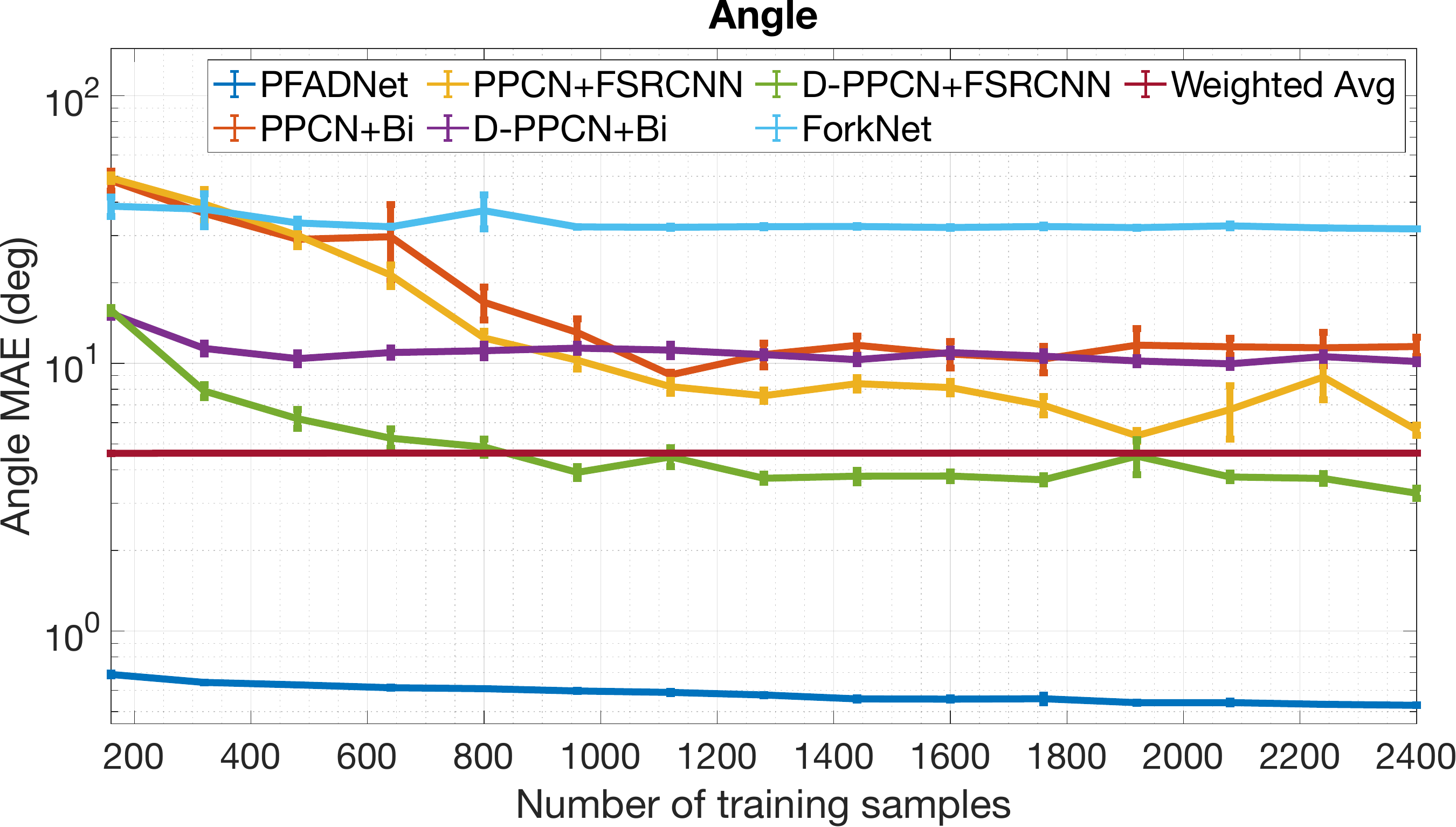}\\
    \includegraphics[width=0.47\linewidth]{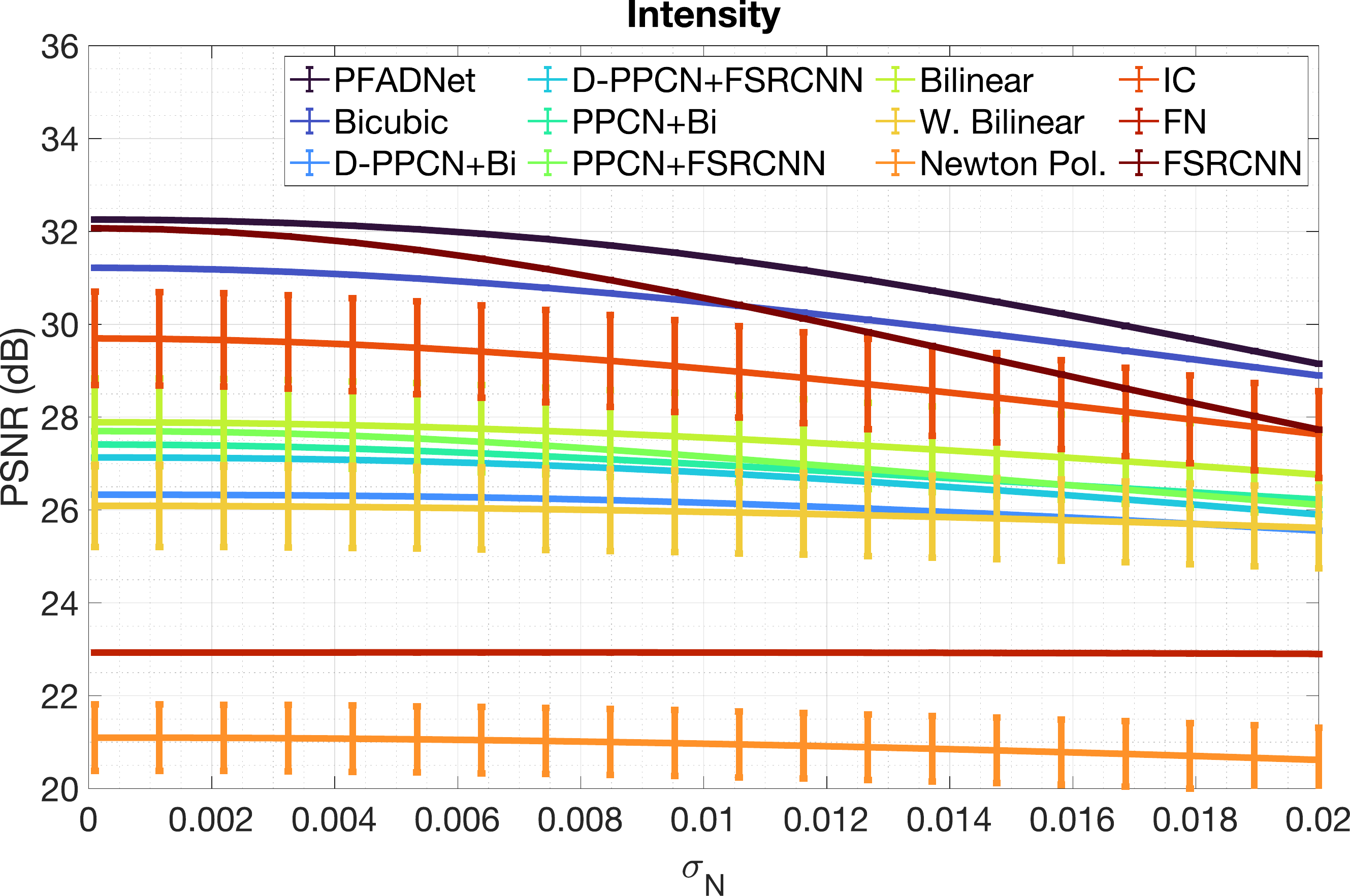}
    \includegraphics[width=0.475\linewidth]{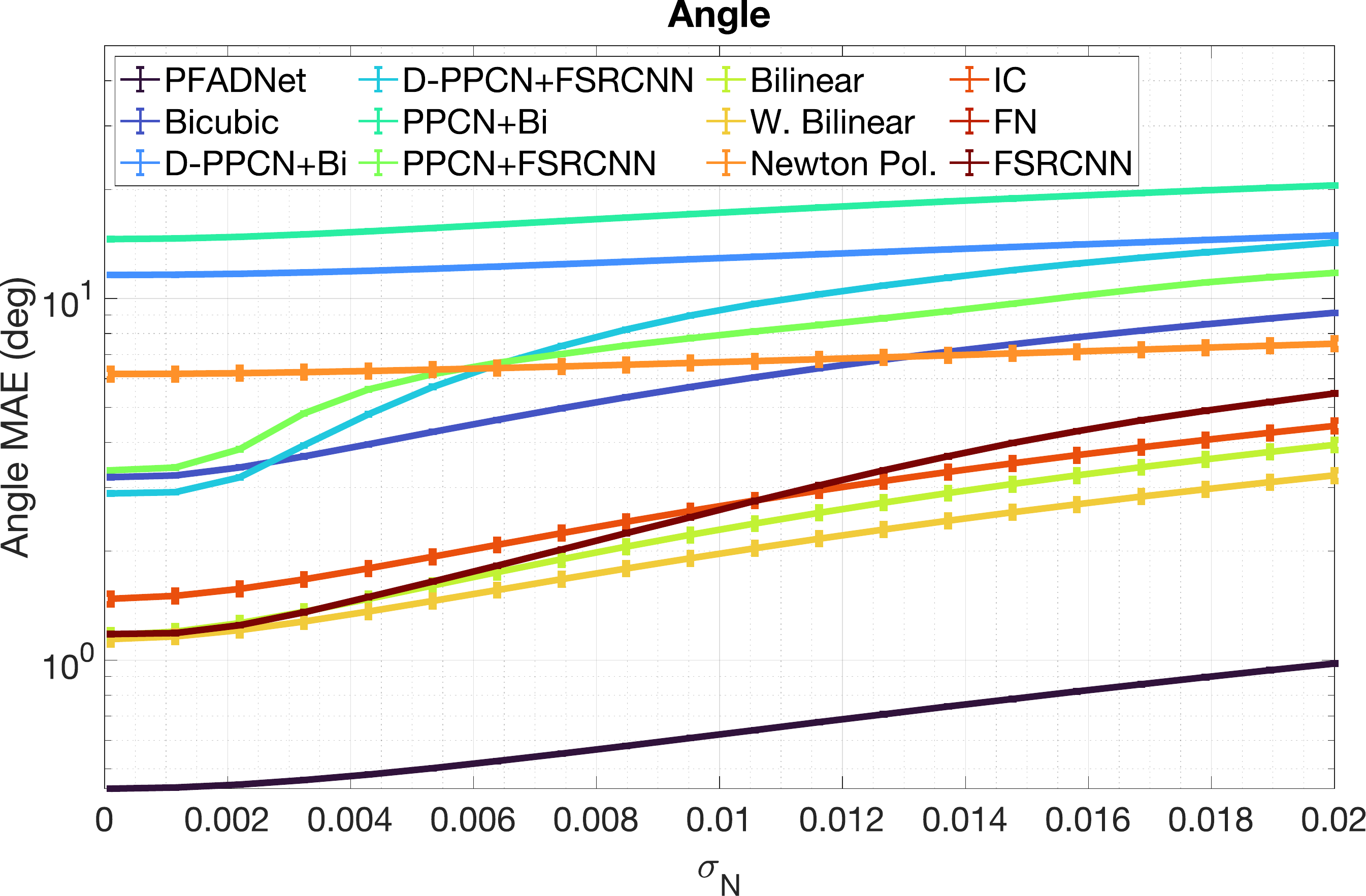}\\
    \caption{Image PSNR and absolute angle error varying the number of training samples (top-row) and additive noise with standard deviaton $\sigma_N$ (bottom-row). Our approach shows a consistent lower angle error and better PSNR when training data comprise at least $250$ samples. Even with additive noise, the angle error of our method remains below $1^\circ$. Note that angle MAE is displayed with logarithmic scale.}
    \label{fig:errorplots}
\end{figure*}

\subsection{Loss Function}

To train our network we propose the Loss function:

\begin{eqnarray}
\mathcal{L}(I,\hat{I},\mathcal{A}, \hat{\mathcal{A}}) &=& \gamma \mathcal{L}_{I} + (1-\gamma) \mathcal{L}_{\mathcal{A}} \nonumber\\
\mathcal{L}_\mathcal{I} &=& (1-\beta) \vert I - \hat{I} \vert + \beta \big(1-\mbox{SSIM}(I,\hat{I}) \big)\nonumber\\
\mathcal{L}_\mathcal{A} &=& \| \mathcal{A} - \hat{\mathcal{A}} \|_2^2
\end{eqnarray}

\noindent defined as a linear combination of the image loss $\mathcal{L}_I$ and
the angle loss $\mathcal{L}_\mathcal{A}$ between the produced angle vector
image $\hat{\mathcal{A}}$ and the ground truth $2$-channel angle image
$\mathcal{A} = ( \cos(2 \Phi), \sin( 2 \Phi))$. The image loss is further
composed by weighting an $L_1$ component with the Structure SIMilarity (SSIM)
function as proposed in~\cite{SSIM2004}. We experimentally observed that good
values for the parameters are $\gamma=0.5$ and $\beta = 0.84$.


\section{Experimental Section}

We compared our PFADN with different state-of-the-art approaches.
In particular, we distinguished between learning and non-learning techniques for PFA demosaicing.

We considered the naive pixel unravelling + bicubic up-scaling as baseline, then we added the bilinear and weighted bilinear interpolation methods as described in \cite{gao2011bilinear, ratliff2009interpolation}.
We also compared with intensity correlation (IC) method~\cite{zhang2016image} and Newton polynomial interpolation \cite{li2019demosaicking}.
Despite their original purpose is image debayering, we also applied two pixel-level fusion methods as described in \cite{kwan2018demosaicing}: \textit{weighted average} and \textit{alpha-trimmed mean filtering} (ATMF).
Such techniques merge all aforementioned results in different ways: alpha-trimmed removes minimum and maximum for each pixel and computes the average of the remaining, while weighted average fusion is data-driven: it weights results from different methods based on mean squared error computed on the training data.
Regarding the learning-based approaches, we compared with FSRCNN, a state-of-the-art super-resolution CNN presented in \cite{dong2016accelerating}.
To have an actual comparison with our method, we independently trained two FSRCNN models: one for intensity images and one for angle values.
We also compared with the CNN-based approach ForkNet~\cite{zeng2019end}, specifically designed for polarimetric demosaicing, and with two versions of PPCN (Polarization Parameter-Constructing Network) proposed in \cite{wang2020end}.
We implemented two PPCN architectures, as described in the original paper: ``PPCN" (4-8-16-8-3 structure, with 347 parameters) and the one we will identify as ``Deep PPCN" (4-96-128-64-32-3 structure, with 23331 parameters).
In both cases, we set the learning rate to $10^{-4}$.
Since PPCN is designed to be a preliminary process for polarimetric tasks, we paired the two PPCN versions with different upscalings: the naive bicubic and the learning-based FSRCNN for super-resolution \cite{dong2016accelerating}.

\subsection{Training}

We initially trained our network (and the competitors) with purely synthetic data. Mosaiced images were generated with intensity images from DIV2K dataset~\cite{Timofte_2018} and random smooth surfaces for the angle. A total of $5120$ different samples were used in this first stage, training the network using the Adam optimizer with a learning rate of $10^{-4}$.
Then, we acquired several image samples (as described in §\ref{sec:calibration}) with a FLIR Blackfly Monochrome Polarisation camera, mounting a Sony IMX250MZR sensor providing PFA mosaiced images of size 2448 x 2048 pixels.
We rotated the camera in $50$ different angles, in each pose we displayed $65$ images (randomly sampled from DIV2K dataset) for a total of $3250$ data samples.
Image samples were divided into training and test sets with ratio 75/25, then the training process was resumed only with real-world data and an adaptive learning rate until convergence.
For a fair comparison, we trained all the learning-based approaches with the same data used for our network, and we used the training procedures as specified in the original papers.

\subsection{Angle and Image Accuracy}

\begin{table}
\begin{center}
\caption{Comparison between different demosaicing approaches.}\label{tab:allcomparisons}
\setlength{\tabcolsep}{0.4em} 
\begin{tabular}{r|c|c|c|}
\cline{2-4}
\multicolumn{1}{l|}{}           & \multicolumn{2}{c|}{$I$}                                    & \multicolumn{1}{c|}{$\Phi$}    \\ \cline{2-4}
\multicolumn{1}{l|}{}           & \multicolumn{1}{l|}{PSNR (dB)} & \multicolumn{1}{l|}{MAE$\times 10^{3}$} & \multicolumn{1}{l|}{MAE $(^\circ)$} \\ \hline \hline
\multicolumn{1}{|r|}{Bicubic}   & $31.21 \pm 5.15$      & $17.15 \pm 21.47$      & $2.89 \pm 8.55$  \\ \hline
\multicolumn{1}{|r|}{Bilinear}  & $28.82 \pm 4.36$      & $20.26 \pm 30.02$     & $1.21 \pm 2.34$  \\ \hline
\multicolumn{1}{|r|}{W. Bilinear}  & $26.96 \pm 4.47$        & $23.50 \pm 38.22$    & $1.18 \pm 2.53$  \\ \hline
\multicolumn{1}{|r|}{Newton Pol.}  & $21.80 \pm 4.56 $      & $34.44 \pm 73.60$     & $6.38 \pm 18.68$  \\ \hline
\multicolumn{1}{|r|}{IC} & $30.68 \pm 5.05$     & $18.12 \pm 22.92$      & $1.52 \pm 1.80$  \\ \hline
\multicolumn{1}{|r|}{Fusion W.Avg} & $30.77 \pm 4.81$     & $18.11 \pm 22.56$   & $3.23 \pm 9.04$        \\ \hline
\multicolumn{1}{|r|}{Fusion ATMF} & $30.58 \pm 4.82$     & $ 18.22 \pm 23.32$      & $2.90 \pm 10.43$  \\ \hline
\multicolumn{1}{|r|}{ForkNet}   & $22.93 \pm 3.15$  & $58.13 \pm 41.38$  & $33.9 \pm 25.9$  \\ \hline
\multicolumn{1}{|r|}{FSRCNN}   & $32.06 \pm 4.48$  & $16.19 \pm 18.93$  & $1.15 \pm 3.38$  \\ \hline
\multicolumn{1}{|r|}{PPCN+Bicubic}     & $27.41 \pm 4.88$     & $28.64  \pm 31.53 $     & $9.26 \pm 14.52$ \\ \hline
\multicolumn{1}{|r|}{D-PPCN+Bicubic}     & $25.55 \pm 5.14$     & $ 35.67 \pm 32.48 $       & $9.77 \pm 13.90$ \\ \hline
\multicolumn{1}{|r|}{PPCN+FSRCNN}     & $29.96 \pm 3.33$     & $27.89 \pm 30.32$         & $5.27 \pm 4.55$ \\ \hline
\multicolumn{1}{|r|}{D-PPCN+FSRCNN}     & $27.47 \pm 3.66$     & $34.37 \pm 27.45 $         & $2.87 \pm 4.97$ \\ \hline
\multicolumn{1}{|r|}{PFADN}     & $\mathbf{32.31 \pm 4.65}$     & $\mathbf{15.91 \pm 19.33}$     & $\mathbf{0.46 \pm 0.39}$ \\ \hline
\end{tabular}\vspace{-0.5cm}
\end{center}
\end{table}

We started comparing the learning-based techniques varying the number of training images.
Since our method is meant to be fine-tuned for each specific camera instance, it is important to know how many training samples to acquire to obtain the desired accuracy.
In all tests we retrained PFADN, PPCN, Deep PPCN (D-PPCN) and ForkNet starting from the initial weights obtained with the purely synthetic dataset.
We used the Adam optimizer with an adaptive learning rate until convergence.
In Figure \ref{fig:errorplots} (top) we display the intensity image PSNR (left) and angle MAE (right) varying the number of training data.
As expected, increasing the number of training samples improves the situation, especially for the intensity.
Our approach achieved a PSNR (Peak Signal-to-Noise Ratio) from 31 (with 200 training samples) to 32 dB, while other techniques are limited to values lower than 28 dB.
The angle error (MAE, Mean Absolute Error) estimated by our PFADN is significantly lower than the one obtained with other techniques: indeed, in all cases PPCN and ForkNet performed consistently worse (note that the y-axis scale is logarithmic).
We guess two reasons for that: first, loss functions used to train PPCN and ForkNet (as reported in the original paper) use the same loss for both the image and the angle.
However, angles wrap with period $\pi$, and this makes very difficult for the network to disambiguate such cases.
Our method, instead, handles the wrapping by doubling the angle and considering the difference between the two corresponding unitary angle vectors.
Second, other networks convolve the mosaiced image with unitary stride, regardless the underlying PFA pattern. Albeit being not optimal, we observed that this tends to oversmooth the image in most the cases.
In the second row of Figure \ref{fig:errorplots} we show image PSNR and angle MAE when increasing an additive noise of input images.
The noise was generated as a zero-mean Gaussian with standard deviation $\sigma_N$ (on x-axis).
In this experiment we compared all the aforementioned polarimetric demosaicing methods: also in this situation PFADN gives better results, suggesting a possible application for polarised image denoising.
In particular, the FSRCNN upscaling exhibited comparable results in terms of image PSNR only for low noise values ($\sigma_N < 0.002$), while for angle error other methods performed quite worse with respect to our PFADN.
Finally, in Table~\ref{tab:allcomparisons} we report image and angle accuracy on the test set against all the considered methods when trained with the full training data.
In terms of intensity image, the upscaling techniques based on bicubic, intensity correlation (IC) and the learning-based FSRCNN offer a result close to our PFADN ($32.31$ dB).
In terms of polarisation angle, some techniques reach an error close to one degree (with some degrees of standard deviation), but overall our approach exhibits an average absolute error of $0.46^\circ$.
Note that we used two separate models for FSRCNN (for intensity and angle), and while the upscaling network performs very well with the intensity image, the angle error is still considerably higher with respect to ours.

\subsection{Qualitative Results}

\begin{figure*}
\begingroup
\setlength{\tabcolsep}{2pt} 
\renewcommand{\arraystretch}{1} 
    \begin{tabular}{cccccc}
        \rotatebox[y=2cm]{90}{GT} &
        \includegraphics[width=0.23\linewidth]{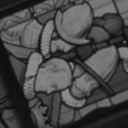} &
        \includegraphics[width=0.23\linewidth]{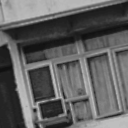} &
        \rotatebox[y=2.0cm]{90}{GT} &
        \includegraphics[width=0.23\linewidth]{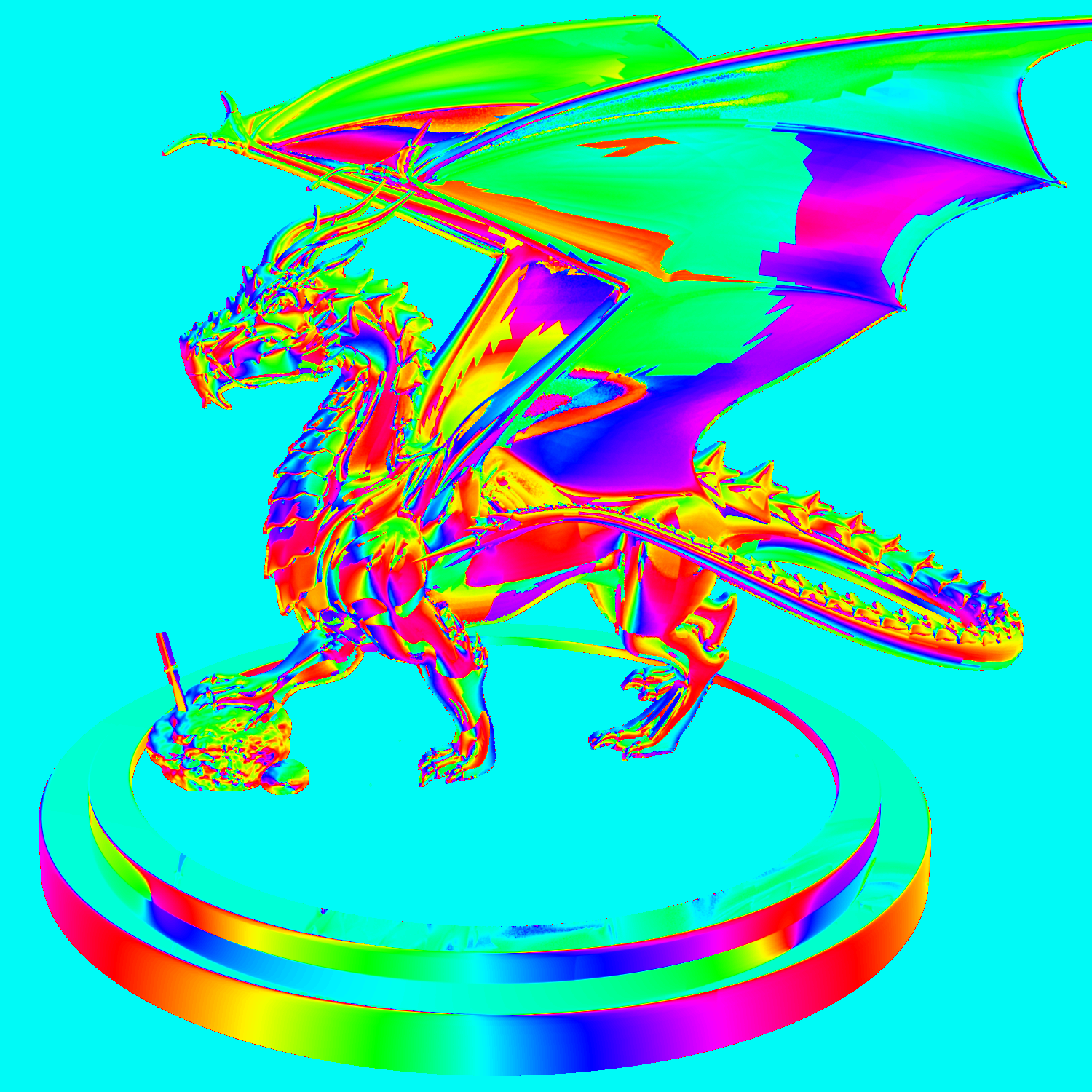} &
        \includegraphics[width=0.23\linewidth]{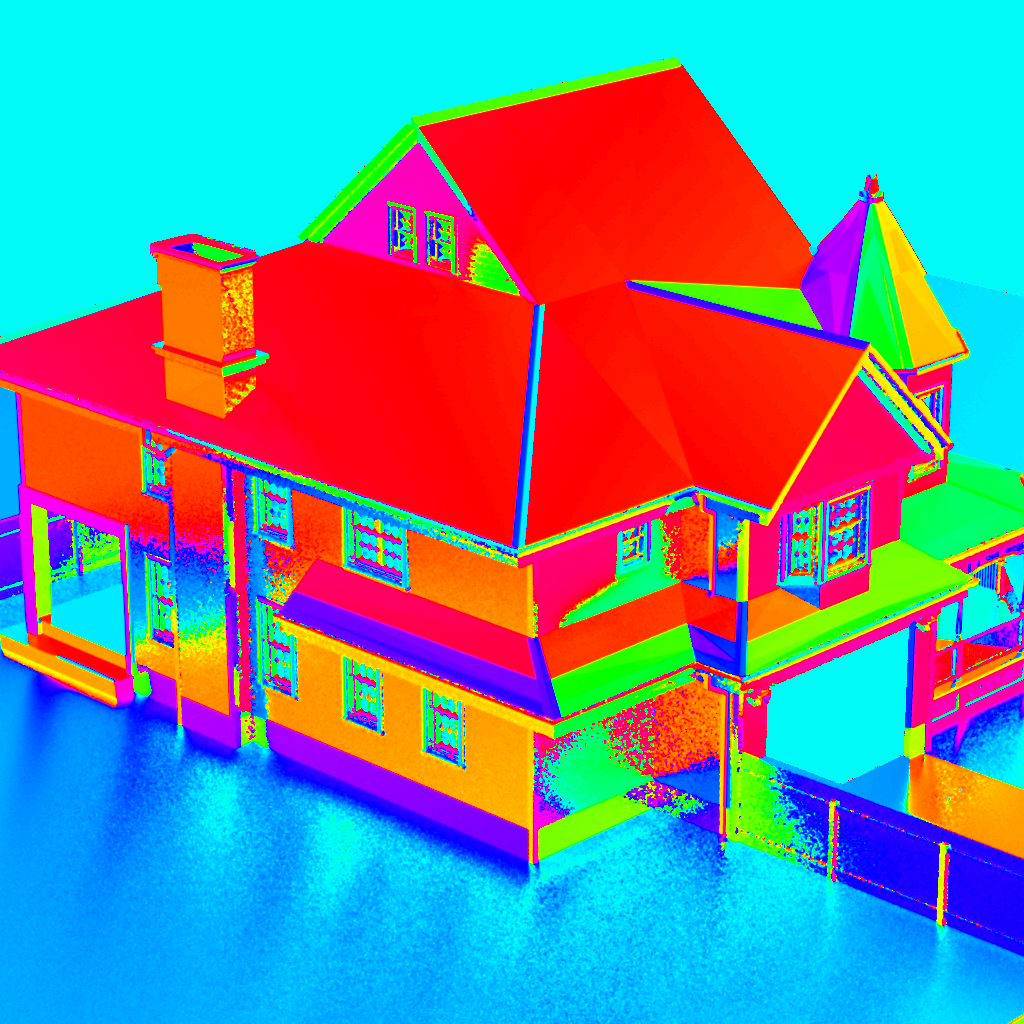} \\
        \rotatebox[y=2cm]{90}{PFADN} &
        \includegraphics[width=0.23\linewidth]{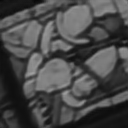} &
        \includegraphics[width=0.23\linewidth]{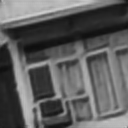} &
        \rotatebox[y=2.0cm]{90}{PFADN} &
        \includegraphics[width=0.23\linewidth]{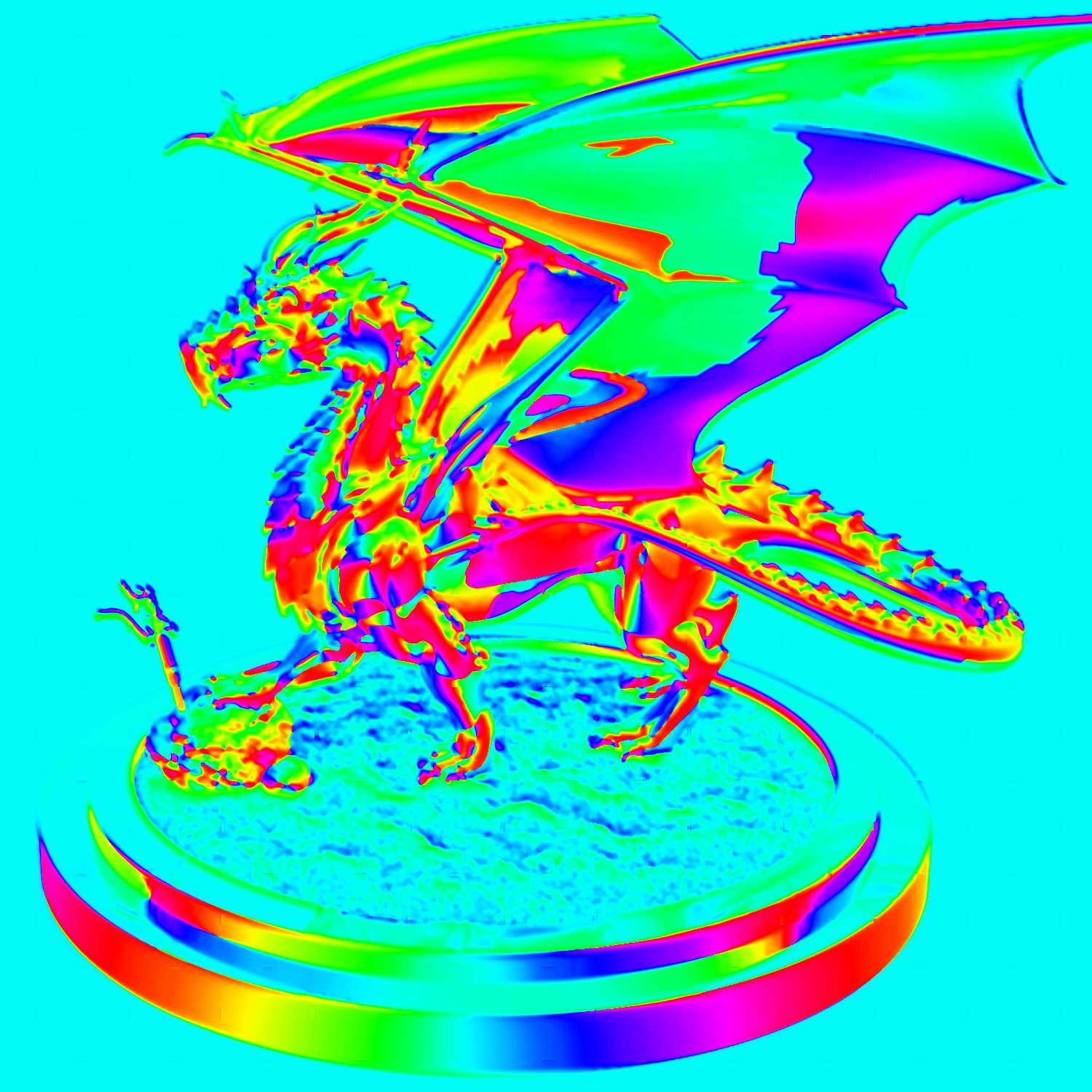} &
        \includegraphics[width=0.23\linewidth]{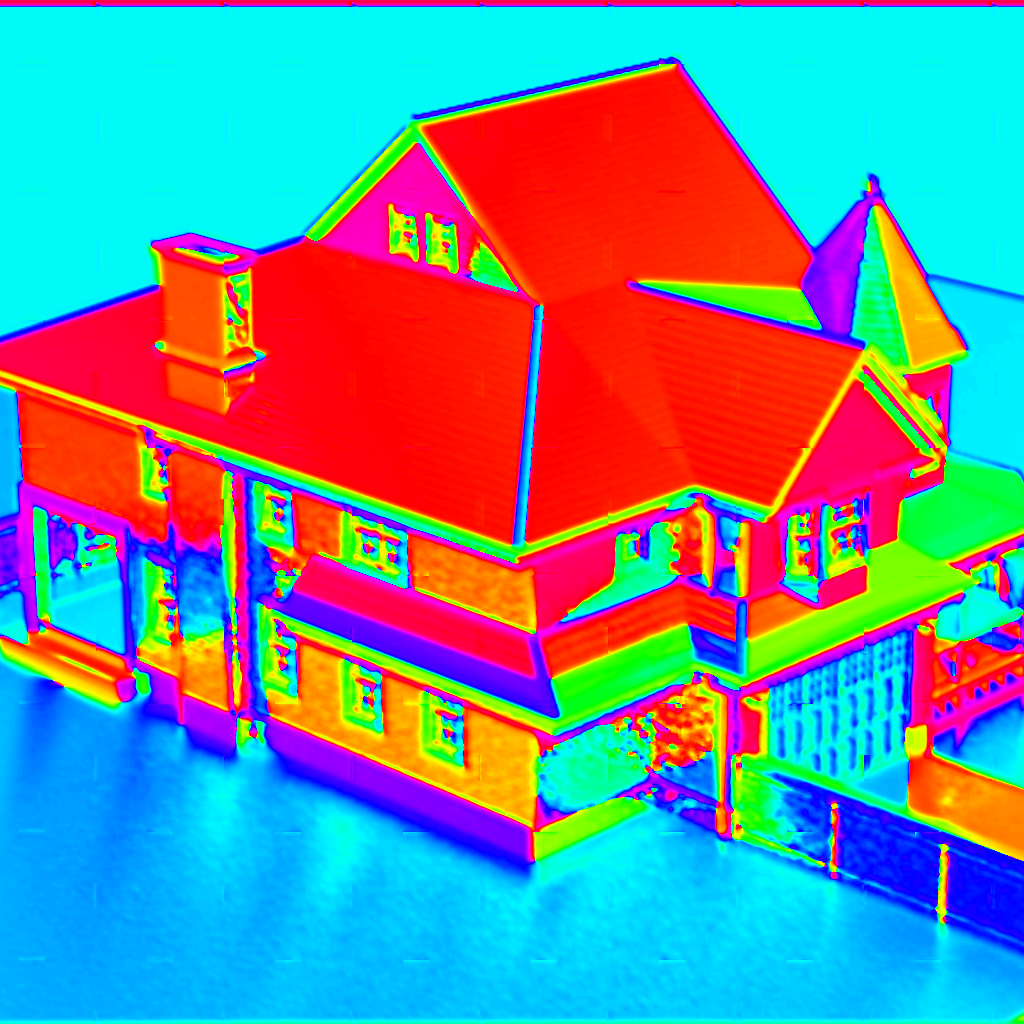} \\
        \rotatebox[y=2cm]{90}{PPCN+FSRCNN} &
        \includegraphics[width=0.23\linewidth]{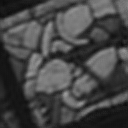} &
        \includegraphics[width=0.23\linewidth]{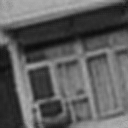} &
        \rotatebox[y=2.0cm]{90}{FSRCNN} &
        \includegraphics[width=0.23\linewidth]{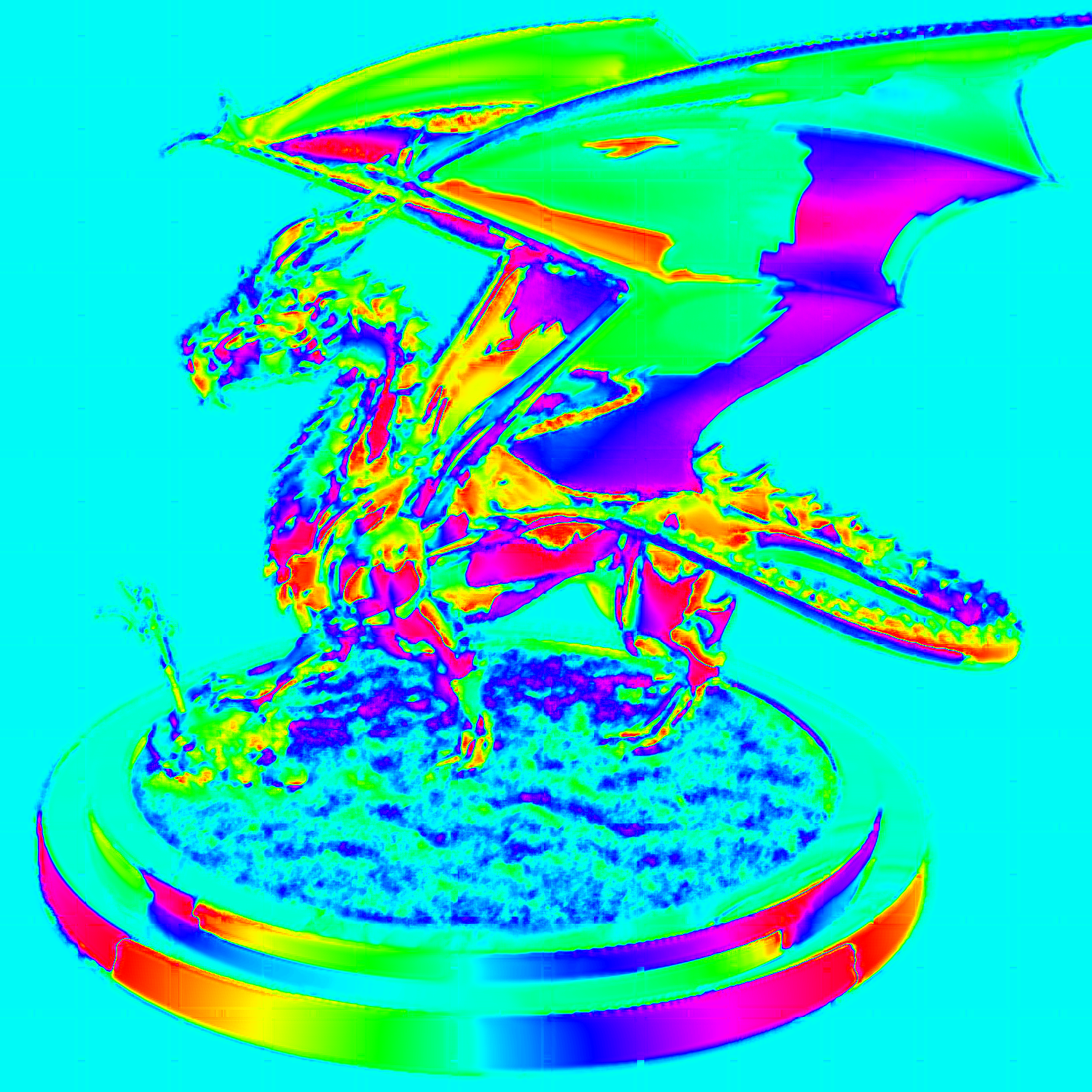} &
        \includegraphics[width=0.23\linewidth]{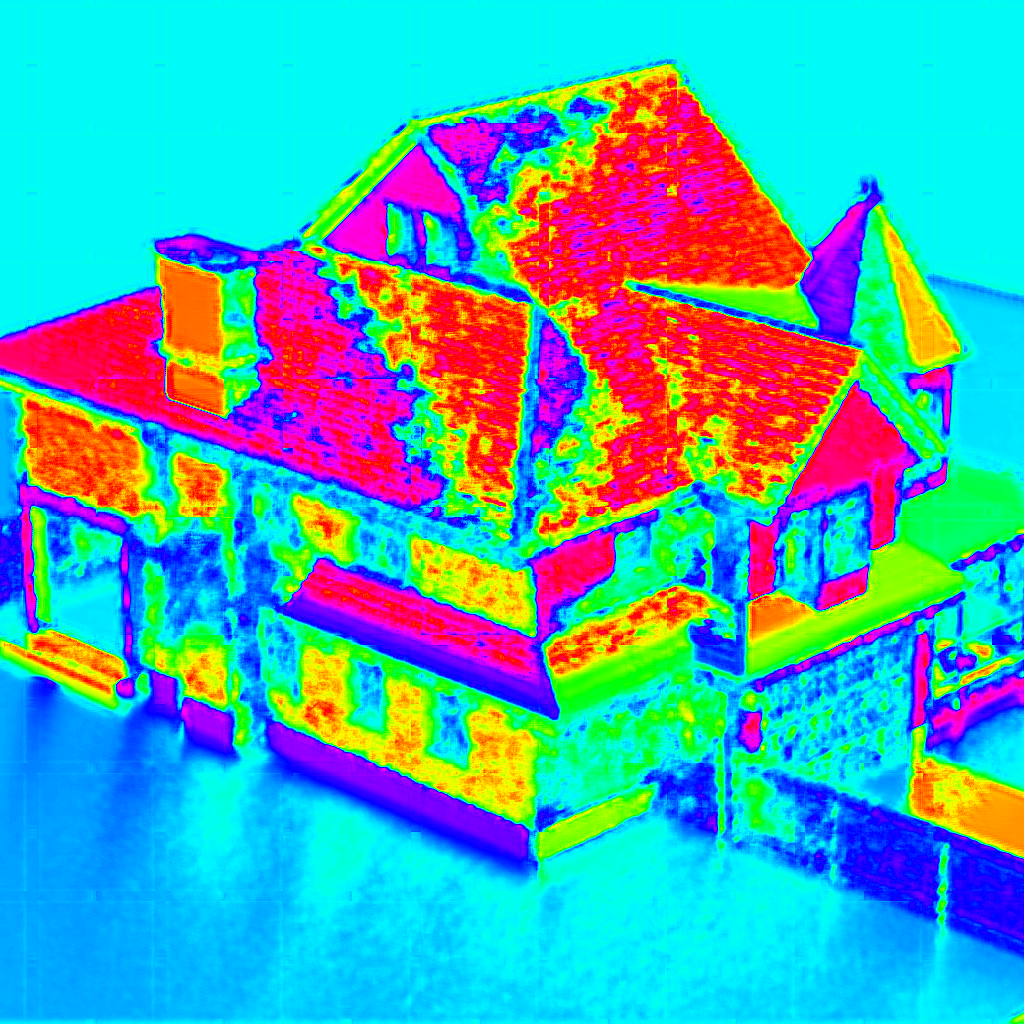} \\
        \rotatebox[y=2cm]{90}{Bicubic} &
        \includegraphics[width=0.23\linewidth]{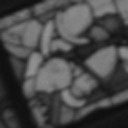} &
        \includegraphics[width=0.23\linewidth]{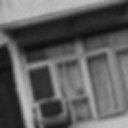} &
        \rotatebox[y=2.0cm]{90}{Bilinear} &
        \includegraphics[width=0.23\linewidth]{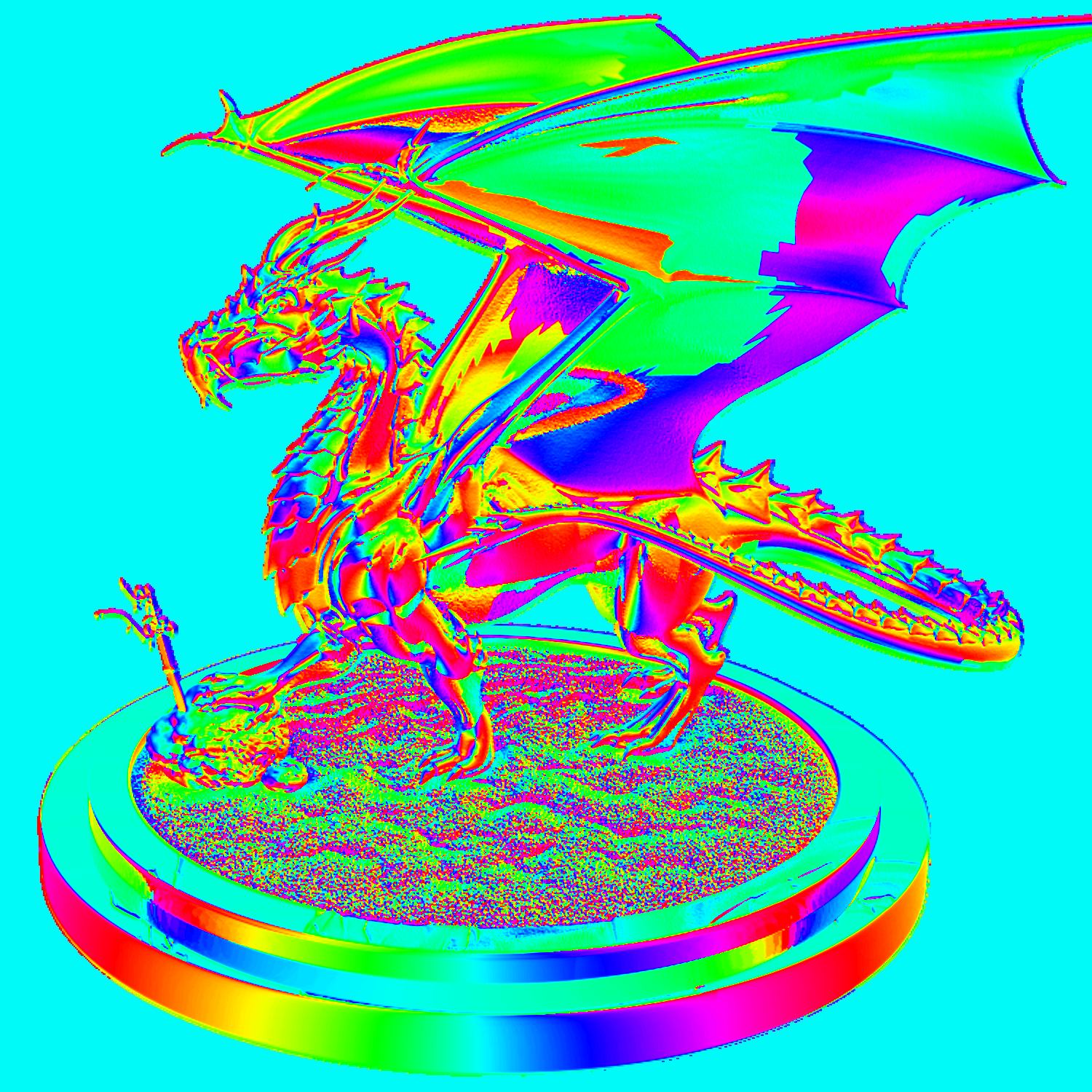} &
        \includegraphics[width=0.23\linewidth]{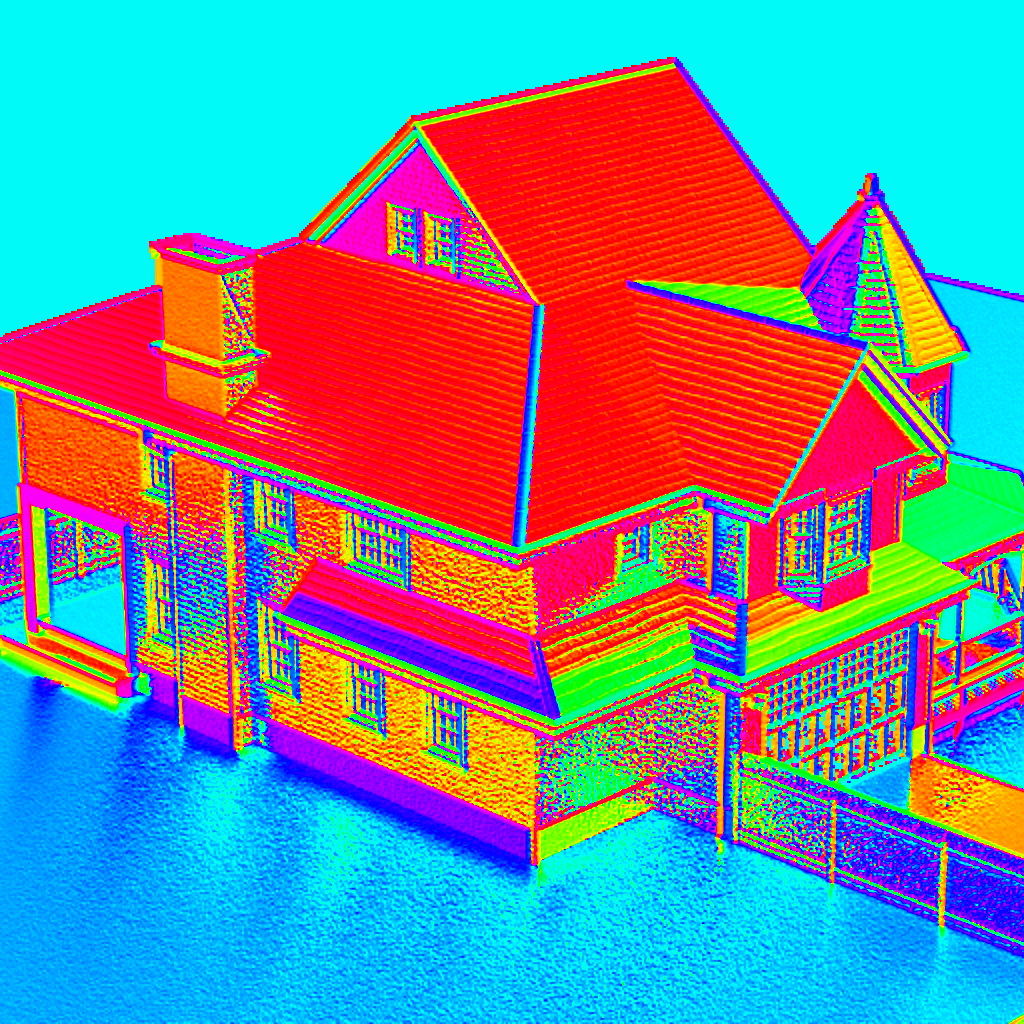} \\
        \rotatebox[y=2cm]{90}{IC} &
        \includegraphics[width=0.23\linewidth]{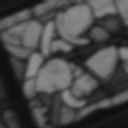} &
        \includegraphics[width=0.23\linewidth]{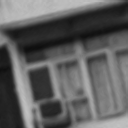} &
        \rotatebox[y=2.0cm]{90}{IC} &
        \includegraphics[width=0.23\linewidth]{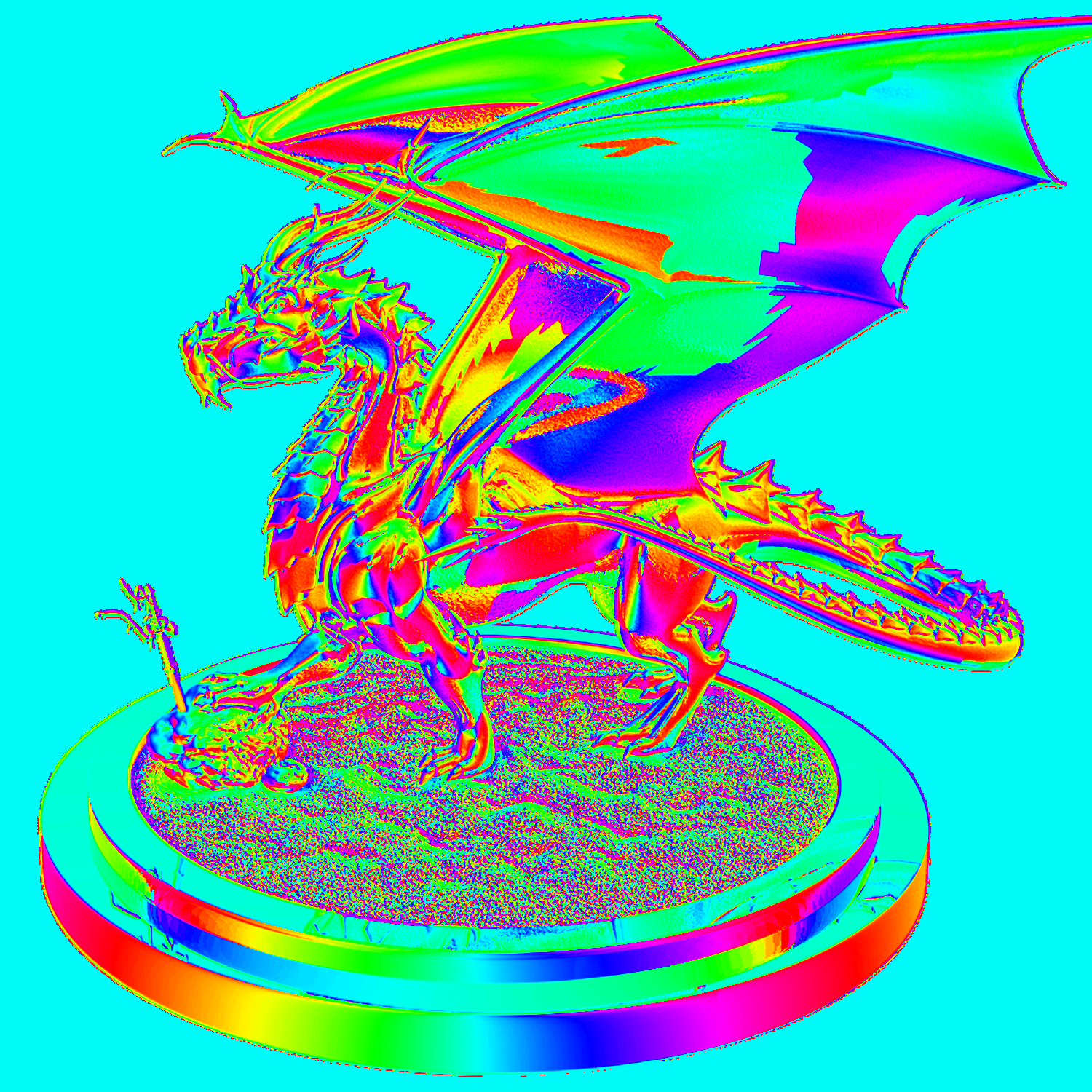} &
        \includegraphics[width=0.23\linewidth]{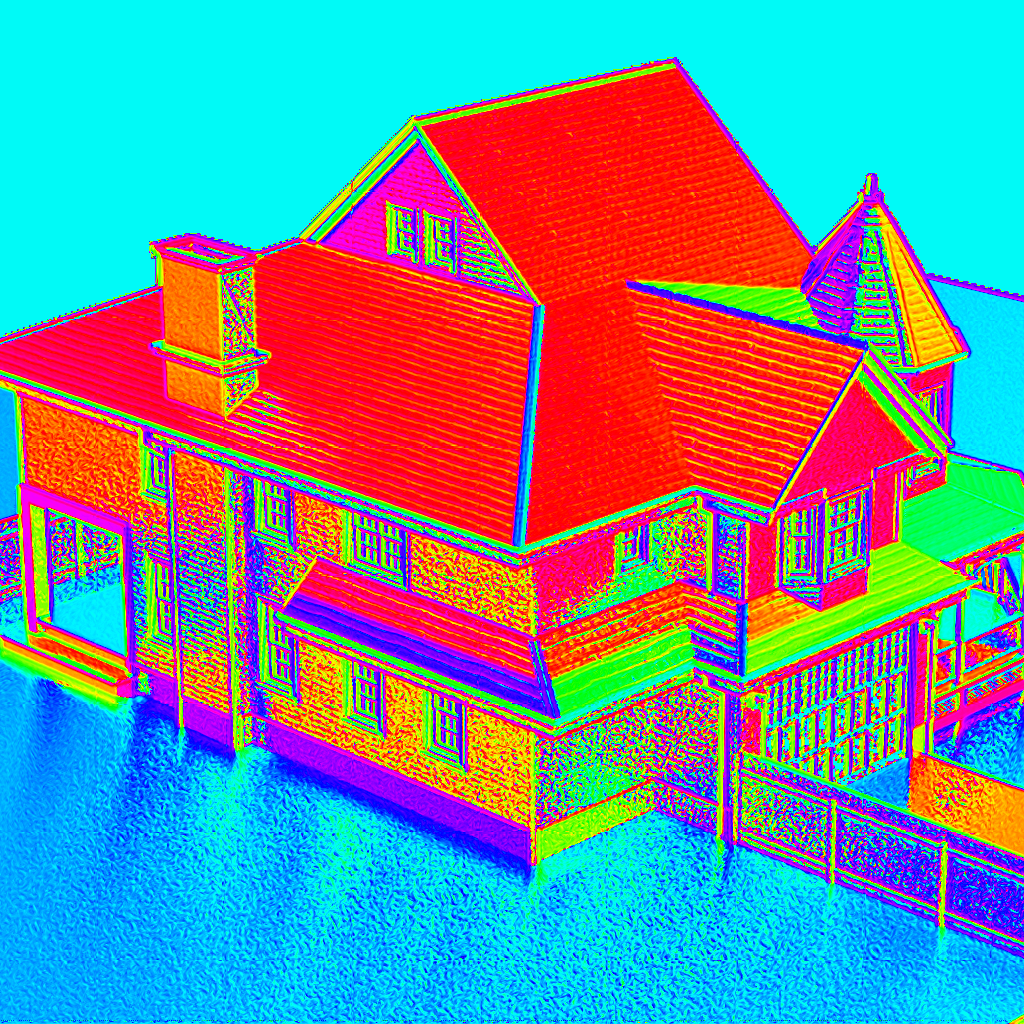} \\
    \end{tabular}
    \endgroup
    \caption{Left: qualitative comparison of demoisaiced intensity $I$ for two samples from the test set. Right: qualitative comparison of AoLP images for two synthetic scenes. In both cases the first row shows ground truth data and the second row the proposed PFADN network output.}
    \label{fig:qualitative_comparison}
\end{figure*}

In the first two columns of Fig.~\ref{fig:qualitative_comparison} we show a qualitative comparison of the demosaiced intensity images $I$ from our test set for different methods.
Overall, PFADN (second row) produces a sharper result closer to the Ground Truth ($1^{st}$ row) in all the cases.
The two ``purely algorithmic" methods IC and Bicubic look blurry at the edges. In this respect, we see no visible advantage in using IC instead of the old but reliable Bicubic interpolation.
The competing Deep Learning based method PPCN+FSRCNN produced a definitely sharper but also noisier result. Overall, the images looks less saturated especially in darker regions.
In order to show some realistic examples of demosaiced AoLP, we synthetically generated some scenes using Mitsuba2 Renderer \cite{nimier2019mitsuba}.
Such tool allows to simulate light polarization state and obtain physically accurate scenes that we composed to generate mosaiced images from a PFA camera.
Of course this approach would discard all the nonlinearities introduced by a real PFA camera that our approach aims to capture, but we still found the qualitative outcomes interesting for evaluation purposes.
The rightmost part of Fig.~\ref{fig:qualitative_comparison} shows AoLP images for two of such scenes: in general our method returns angles that are less noisy and closer to the respective ground truth.

\begin{figure*}[t]
    \includegraphics[width=0.99\linewidth]{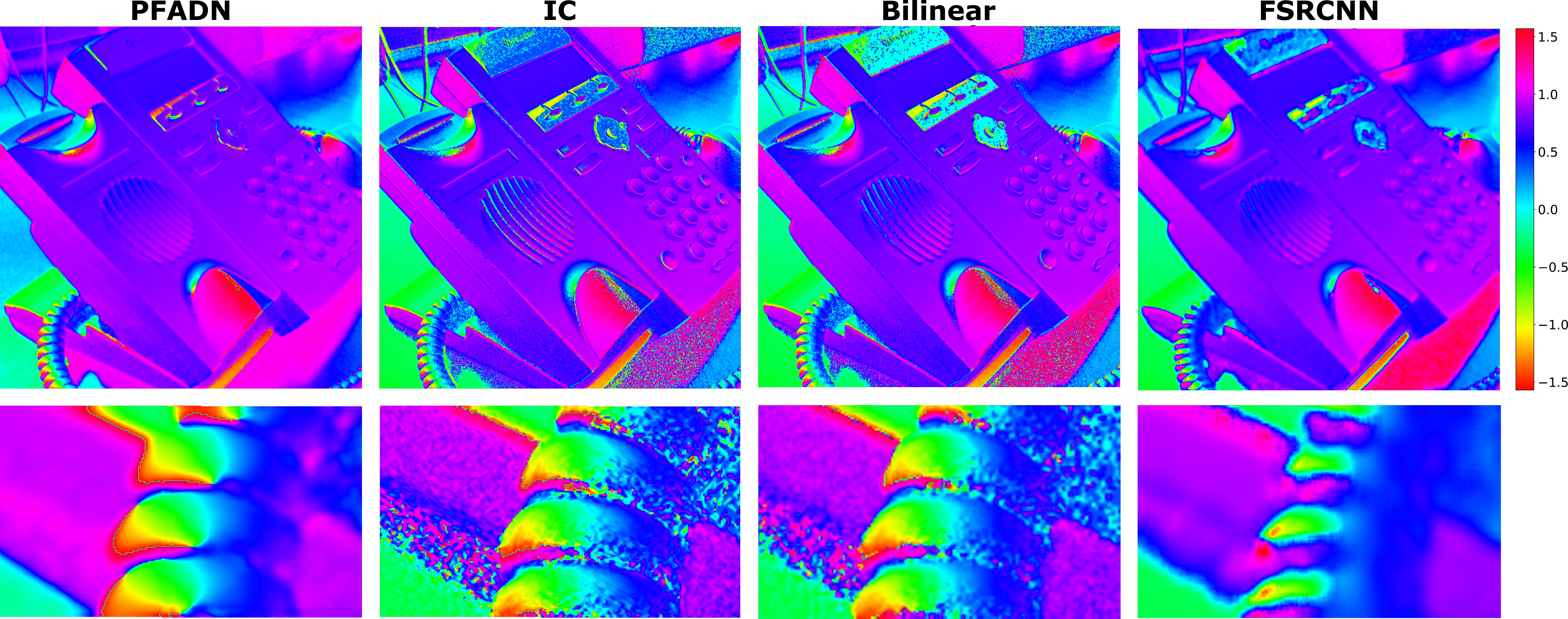}
    \caption{Demosaiced AoLP (radians) of a real scene with different methods (our PFADN, Intensity Correlation, bilinear interpolation and FSRCNN).}
    \label{fig:real_image}
\end{figure*}

Finally, in Fig.~\ref{fig:real_image} we show the resulting angle $\Phi$ of a real object acquired with our camera and demosaiced with four methods.
Our method (first column) clearly exhibits a less noisy behaviour (second rows show a detail of the top image), while keeping sharp edges between differently polarised areas. Differences in the background and in regions around the display are due to a DOLP $\approx 0$ making the angle undefined in practice.

\section{Conclusions}

We proposed a CNN-based camera model to demosaic a PFA camera image into a full-resolution intensity image and AoLP.
The network is trained with real-world data acquired using a consumer LCD screen, with a technique independent by monitor gamma and other non-linearities produced by the tilted camera-screen configuration. In this way, the model can be fine-tuned for a specific camera instance to produce results exceeding the current state-of-the-art of both learning and algorithmic approaches without requiring expensive setups.
The novelty of our network architecture resides in the Mosaiced Convolution operation, taking advantage (and preserving) the underlying orientation pattern of the PFA.
In the future, we aim to generalize the Mosaiced Convolution blocks to consider arbitrary kernel shapes and hence increasing the combination of repeated orientation patterns.


%




\ifCLASSOPTIONcaptionsoff
  \newpage
\fi



\bibliographystyle{IEEEtran}
\bibliography{bibliography}
%

%

\begin{IEEEbiographynophoto}{Mara Pistellato}
received the PhD degree in Computer Science from University Ca'Foscari of Venice in 2020, where she is a postdoc researcher. Her interests are in the areas of Computer Vision and Pattern Recognition, in particular 3D reconstruction techniques for real-world applications, machine learning for accurate surface reconstruction and polarimetric imaging.
\end{IEEEbiographynophoto}

\begin{IEEEbiographynophoto}{Filippo Bergamasco}
received the PhD degree in Computer Science from University Ca'Foscari Venice, Italy, in 2015. He is currently an Assistant Professor at Ca'Foscari University of Venice. His research interests are in the area of computer vision and machine learning, ranging from 3D reconstruction, camera calibration to structure from motion, structured-light scanning, and remote sensing.
\end{IEEEbiographynophoto}

\begin{IEEEbiographynophoto}{Tehreem Fatima}
Is currently a PhD student at Ca'Foscari University of Venice. Her research interests are in the area of pattern recognition and machine learning for computer vision tasks, in particular, polarimetric camera calibration and its applications.
\end{IEEEbiographynophoto}

\begin{IEEEbiographynophoto}{Andrea Torsello}
received his PhD in computer science at the University of York, UK. From 2007 he is with Ca'Foscari University of Venice, Italy, where he is Full Professor. His research interests are in the areas of Computer Vision and Pattern Recognition, in particular the interplay between Stochastic and Structural approaches as well as Game-Theoretic and Physical models.
\end{IEEEbiographynophoto}



\end{document}